\documentclass[runningheads]{llncs}
\usepackage{graphicx}

\usepackage{tikz}
\usepackage{comment}
\usepackage{amsmath,amssymb} 
\usepackage{color}
\usepackage{colortbl}
\usepackage{subcaption}
\usepackage{wrapfig}
\usepackage{enumitem}
\usepackage{multirow}
\usepackage{pifont}

\usepackage{arydshln}
\usepackage{bm}

\newcommand{\cmark}{\ding{51}}%
\newcommand{\xmark}{\ding{55}}%
\newcommand{\boldOrange}[1]{\textcolor{orange}{\textbf{#1}}}
\newcommand{\boldBlue}[1]{\textcolor{blue}{\textbf{#1}}}

\newcommand{\boldPurple}[1]{\textcolor{purple}{\textbf{#1}}}

\usepackage{array} 
\newcolumntype{x}[1]{>{\centering\arraybackslash}p{#1pt}}

\newlength\savewidth\newcommand\shline{\noalign{\global\savewidth\arrayrulewidth
		\global\arrayrulewidth 1pt}\hline\noalign{\global\arrayrulewidth\savewidth}}

\newcommand{\ie}{\textit{i}.\textit{e}.}
\newcommand{\eg}{\textit{e}.\textit{g}.}

\usepackage[pagebackref,breaklinks,colorlinks]{hyperref}

\definecolor{mygray}{gray}{0.5}

\begin{document}

\pagestyle{headings}
\mainmatter
\def\ECCVSubNumber{434}  



\title{LocVTP: Video-Text Pre-training for\\Temporal Localization}

\titlerunning{LocVTP}
%
\author{Meng Cao\inst{1} \and
Tianyu Yang\inst{2} \and
Junwu Weng\inst{2}\and 
Can Zhang\inst{1} \and
Jue Wang\inst{2} \and
Yuexian Zou\inst{1,3}}
\authorrunning{M.Cao et al.}
%
\institute{School of Electronic and Computer Engineering, Peking University \and
Tencent AI Lab \and Peng Cheng Laboratory 
}
\maketitle
\vspace{-1em}
\begin{abstract}
Video-Text Pre-training (VTP) aims to learn transferable representations for various downstream tasks from large-scale web videos. To date, almost all existing VTP methods are limited to \emph{retrieval-based} downstream tasks, \eg, video retrieval, whereas their transfer potentials on \emph{localization-based} tasks, \eg, temporal grounding, are under-explored. In this paper, we experimentally analyze and demonstrate the incompatibility of current VTP methods with localization tasks, and propose a novel \textbf{Loc}alization-oriented \textbf{V}ideo-\textbf{T}ext \textbf{P}re-training framework, dubbed as \textbf{LocVTP}. Specifically, we perform the fine-grained contrastive alignment as a complement to the coarse-grained one by a clip-word correspondence discovery scheme. To further enhance the temporal reasoning ability of the learned feature, we propose a context projection head and a temporal aware contrastive loss to perceive the contextual relationships. Extensive experiments on four downstream tasks across six datasets demonstrate that our LocVTP achieves state-of-the-art performance on both retrieval-based and localization-based tasks. Furthermore, we conduct comprehensive ablation studies and thorough analyses to explore the optimum model designs and training strategies. Codes are available at \url{https://github.com/mengcaopku/LocVTP}.
\end{abstract}
\vspace{-2em}
\section{Introduction}
\begin{figure}[t]
     \centering
     \begin{subfigure}[b]{0.5\textwidth}
         \centering
         \includegraphics[width=\textwidth]{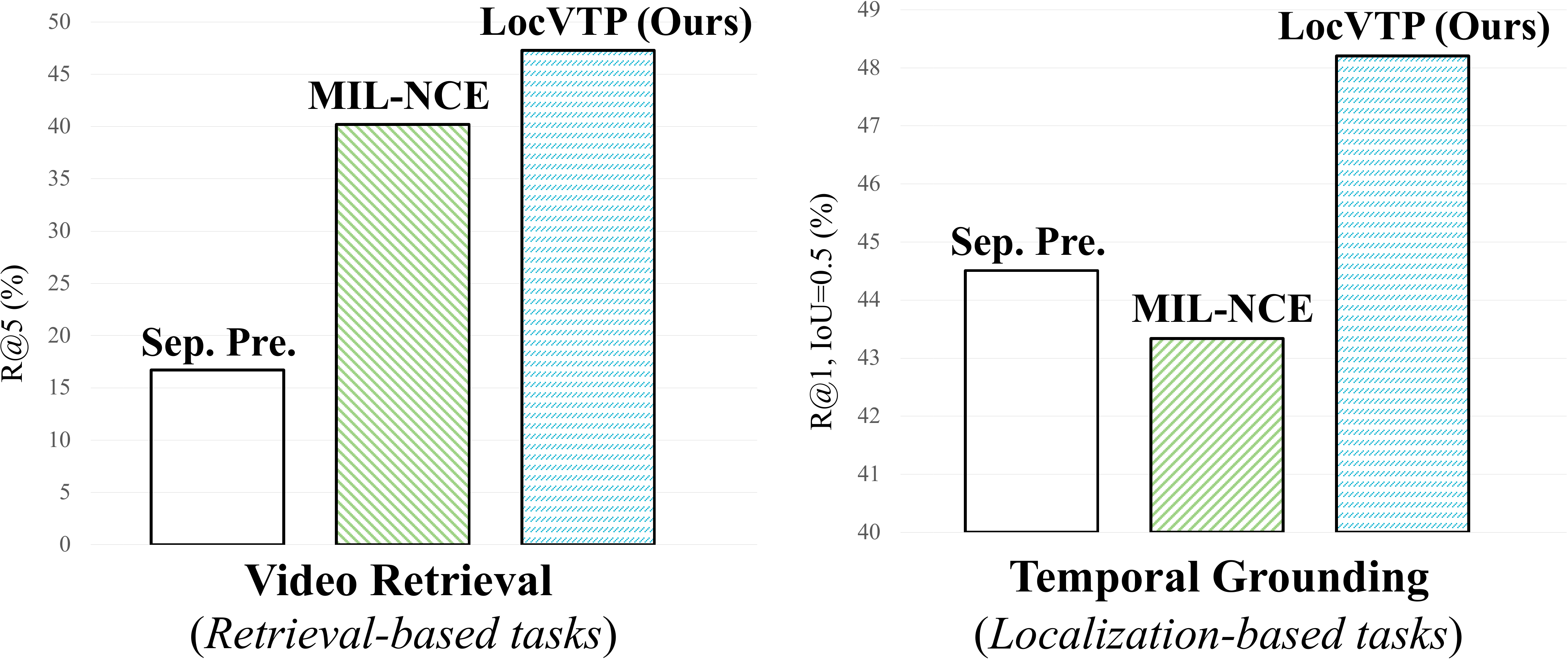}
         \caption{}
         \label{fig:teaserBar}
     \end{subfigure}
     \hfill
     \begin{subfigure}[b]{0.46\textwidth}
         \centering
           \includegraphics[width=0.95\textwidth]{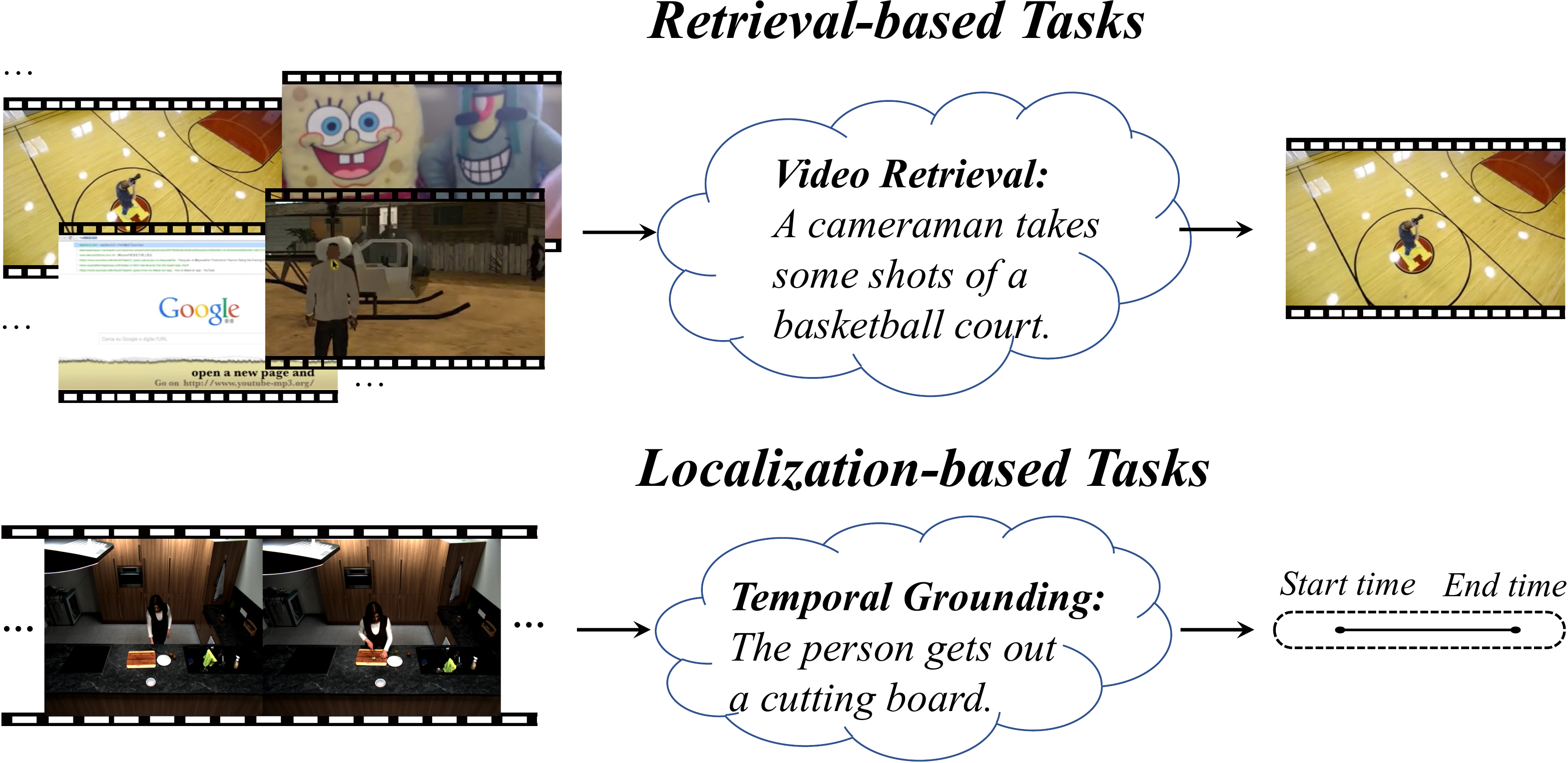}
         \caption{}
         \label{fig:retrievalVSground}
     \end{subfigure}
     \vspace{-2mm}
     \caption{\footnotesize{(a) \textbf{Video retrieval and temporal grounding performance using different pre-trained features.} \texttt{Sep.}\texttt{Pre.} means separately pre-training, \ie, the video encoder supervisedly pre-trained on Kinetics~\cite{kay2017kinetics} and text encoder taken from BERT~\cite{devlin2018bert}. \texttt{MIL-NCE} and our \texttt{LocVTP} are VTP methods pre-trained on HowTo100M~\cite{miech2019howto100m}. For video retrieval, we use COOT~\cite{ging2020coot} as the downstream method and evaluate on YouCook2~\cite{zhou2018towards} dataset with R@5. For temporal grounding, we take 2D-TAN~\cite{zhang2020learning} as the downstream method and evaluate on ActivityNet Captions~\cite{krishna2017dense} dataset with R@1, IoU=0.5. \textbf{(b) Retrieval-based and localization-based downstream tasks.} We take video retrieval and temporal grounding as typical examples, respectively. The former needs video-level classification while the latter requires clip-level or frame-level localization.}}
     \vspace{-3mm}
\end{figure}
Video-Text Pre-training (VTP)~\cite{sun2019videobert,miech2019howto100m,liu2019use,liu2021hit,lei2021less,bain2021frozen,yan2021video,wang2021object} has attracted increasing attention with the aim to learn generic and transferable \emph{joint} video-language (VL) representations. Compared to the conventional \emph{separate} pre-training on each single modality, \eg, video features are pre-trained under the action recognition datasets (Kinetics~\cite{kay2017kinetics}, Sport1M~\cite{KarpathyCVPR14}), VTP has several advantages: 1) It leverages large-scale unlabeled narrated video data with automatically generated corresponding text data for video-text correspondence pre-training. 2) It tries to map different modality features into a shared latent space, which reduces the difficulties of the cross-modal feature interaction. Thanks to these advantages, VTP has significantly improved the performance of many downstream VL tasks. For example, as illustrated in \cite{ging2020coot}, the video retrieval performance using features pre-trained with the VTP method \texttt{MIL-NCE}~\cite{miech2020end} is much higher than that using separately pre-trained way (cf. Fig.~\ref{fig:teaserBar} (left)).

Despite their encouraging performance, we find that most current VTP methods are applicable to limited downstream tasks, \ie, they focus on \emph{retrieval-based} tasks which require video-level predictions, \eg, video retrieval~\cite{xu2016msr}, video captioning~\cite{rohrbach2015dataset}, and video question answering~\cite{jang2017tgif}. In contrast, there exists another mainstream \emph{localization-based} tasks which expect more fine-grained clip-level or frame-level predictions, \eg, temporal grounding~\cite{gao2017tall}, action segmentation~\cite{tang2019coin}, action step localization~\cite{zhukov2019cross} (cf. Fig.~\ref{fig:retrievalVSground}). Unfortunately, through experiments, we find their poor generalization abilities on this type of downstream tasks. For example, on temporal grounding, even pre-trained with a much larger dataset HowTo100M~\cite{miech2019howto100m}, the VTP method \texttt{MIL-NCE} still performs worse than the separately pre-trained counterpart (cf. Fig~\ref{fig:teaserBar} (right)).

\begin{figure}[t]
	\centering
	\includegraphics[width=0.95\textwidth]{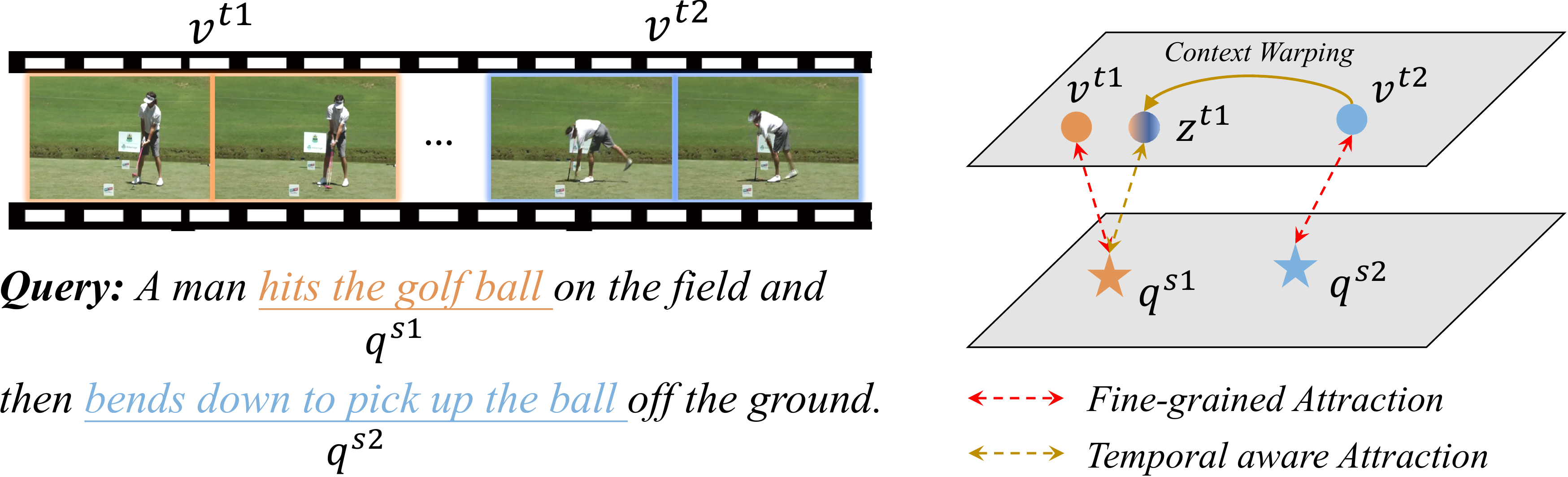}
	\vspace{-2mm}
	\caption{\footnotesize{\textbf{Fine-grained video-text alignment:} positive clip-word pairs are selected via cosine similarity and are then forced to be close to each other, \ie, $\boldsymbol{v}^{t1} \leftrightarrow \boldsymbol{q}^{s1}$, $\boldsymbol{v}^{t2} \leftrightarrow \boldsymbol{q}^{s2}$; \textbf{Temporal relation reasoning:} a context warping head reconstructs $\boldsymbol{v}^{t1}$ conditioned on $\boldsymbol{v}^{t2}$ and distance $t2 - t1$ while maintaining the cross-modal alignment unchanged, \ie, $\boldsymbol{z}^{t1} = \operatorname{warp}(\boldsymbol{v}^{t2}, t2-t1) \leftrightarrow \boldsymbol{q}^{s1}$.}}
	\label{fig:TeaserMotivate}
	\vspace{-4mm}
\end{figure}

In this paper, we analyze that this poor transfer ability on localization-based tasks is due to the absence of two indispensable characteristics: \textbf{1) \emph{Fine-grained alignment}}: We contend that the alignment should be conducted on more \emph{fine-grained} clip-word level instead of the \emph{coarse-grained} video-sentence\footnote{Here we use ``sentence" to represent the whole paired text for each video, such as the ASR in HowTo100M~\cite{miech2019howto100m} or query language in ActivityNet Caption~\cite{krishna2017dense}.} level. As the temporal grounding example shown in Fig.~\ref{fig:TeaserMotivate}, a given query sentence may contain multiple actions (\eg, ``\texttt{hit the golf ball}" ($q^{s1}$) and ``\texttt{bend down to pick up the ball}" ($q^{s2}$)). Thus, aligning each action (or words) to the corresponding clips (\ie, $v^{t1}$ and $v^{t2}$) will help to obtain more detailed and accurate feature representations. \textbf{2) \emph{Temporal relation reasoning}}: We hope the clip features of a certain action can also perceive other actions in the same video. For example, for a typical \texttt{golf} video, action $q^{s2}$ (``\texttt{bend down to pick up the ball}") always occurs shortly after action $q^{s1}$ (``\texttt{hit the golf ball}"). Thus, incorporating such temporal relationship into VTP can help to improve the temporal awareness of video features.

Based on these observations, we propose a novel video-text pre-training framework for localization tasks, dubbed as LocVTP. By considering both above-mentioned characteristics, LocVTP achieves state-of-the-art performance not only on the widely studied retrieval-based tasks, but also on the less-focused localization-based tasks. Specifically, \textbf{for fine-grained alignment}, we extend the coarse-grained contrastive training with video-sentence alignment to a fine-grained one with clip-word alignment. Since there are no clip-word correspondence annotations in existing large-scale datasets, we utilize the latent space established by the coarse-grained contrastive learning to estimate the clip-word similarity, and then select the clip-word pairs with high similarities as positive samples. To further illustrate this, as shown in Fig.~\ref{fig:TeaserMotivate} (right), suppose $\{\boldsymbol{v}^{t1}, \boldsymbol{q}^{s1}\}$ and $\{\boldsymbol{v}^{t2}, \boldsymbol{q}^{s2}\}$ are two matched clip-word feature pairs. Semantic embeddings in each pair are mapped to be close to each other, \ie, $\boldsymbol{v}^{t1} \leftrightarrow \boldsymbol{q}^{s1}$, $\boldsymbol{v}^{t2} \leftrightarrow \boldsymbol{q}^{s2}$. \textbf{For temporal relation reasoning}, we propose a new pretext task called \emph{context warping}. Here we use Fig.~\ref{fig:TeaserMotivate} (right) for illustration. Context warping is designed to generate a new temporally relevant clip features $\boldsymbol{z}^{t1}$, which imitates $\boldsymbol{v}^{t1}$, conditioned on another clip $\boldsymbol{v}^{t2}$ and the relative distance $t2 - t1$ in time, \ie, $\boldsymbol{z}^{t1}=\operatorname{warp}(\boldsymbol{v}^{t2}, t2 - t1)$. The predicted relevant clip feature $\boldsymbol{z}^{t1}$ is enforced to maintain the original established cross-modal correspondence unchanged, \ie, $\boldsymbol{z}^{t1} \leftrightarrow \boldsymbol{q}^{s1}$. In this manner, we simulate the contextual reasoning process and enhance the temporal awareness of video features.

We conduct extensive experiments on four downstream tasks (\ie, video retrieval, temporal grounding, action step localization, and action segmentation) across six datasets. The results on both retrieval-based and localization-based tasks demonstrate the superiority and the generalization ability of our LocVTP. 

In summary, we make three contributions in this paper:

\begin{itemize}[topsep=0pt, partopsep=0pt, leftmargin=13pt, parsep=0pt, itemsep=3pt]
    \item We propose a localization-oriented video-text pre-training framework, LocVTP, which benefits both retrieval-based and the less-explored localization-based downstream tasks.
    \item We pinpoint two crucial designs in LocVTP, \ie, fine-grained video-text alignment and temporal relation reasoning.
    \item Experimental results show that our LocVTP significantly outperforms previous state-of-the-art methods when transferred to various downstream tasks. 
\end{itemize}
\section{Related Work}\label{relatedwork}

\noindent \textbf{Video-Text Pre-training (VTP).} With the release of the large-scale instructional dataset HowTo100M, VTP has spurred significant interest in the community. Overall, the mainstream methods can be broadly classified into two classes: 1) Generative methods: Several methods~\cite{li2019visualbert,lu2019vilbert,tan2019lxmert,chen2020uniter,hu2021transformer,wang2021t2vlad,liu2019use,wang2021dig} try to extend BERT~\cite{vaswani2017attention} to the cross-modal domain, \ie, they accept both visual and textual tokens as input and perform the masked-token prediction task. 2) Discriminative methods. These methods~\cite{lei2021less,bain2021frozen,patrick2020support,liu2021hit} learn representations by differentiating input samples using objectives such as the metric loss~\cite{hoffer2015deep,wu2017sampling} or contrastive loss~\cite{he2020momentum,chen2020simple}. ClipBert~\cite{lei2021less} enables affordable pre-training from sparsely sampled frames. Frozen~\cite{bain2021frozen} adapts the recent ViT~\cite{dosovitskiy2020image} as the visual encoder and is flexible to be trained on both image and video datasets. T2VLAD~\cite{wang2021t2vlad} and FCA~\cite{han2021fine} also perform the fine-grained interactions between video clips and phrases. However, both of them resort to additional overload, \eg, k-means cluster or graph auto-encoder. In contrast, our LocVTP explicitly models the clip-word matching with a more light-weighted similarity comparison manner.

\noindent \textbf{Pre-training for localization tasks.} Compared to the retrieval tasks~\cite{xu2016msr,rohrbach2015dataset,jang2017tgif} which only require only video-level predictions, localization tasks~\cite{gao2017tall,tang2019coin,zhukov2019cross} are essentially different since they need dense clip-level or frame-level predictions and thus the pre-training for these tasks is more challenging. In the pure video domain, this gap has been noticed and several pre-training works~\cite{xu2021boundary,alwassel2021tsp,xu2021low,zhang2022unsupervised} tailored for action localization have been proposed. BSP~\cite{xu2021boundary} synthesizes temporal boundaries using existing action recognition datasets and conducts boundary type classification to generate localization-friendly features. TSP~\cite{alwassel2021tsp} trains video encoders to be temporally sensitive by predicting the foreground clip label and classifying whether a clip is inside or outside the action. As for the video-language domain, our LocVTP is the first pre-training framework designed for localization tasks. Besides, compared to TSP and BSP which require label information for supervised pre-training, our LocVTP can directly learn from narrated videos.
\section{Approach}
\begin{figure}[t]
	\centering
	\includegraphics[width=0.9\textwidth]{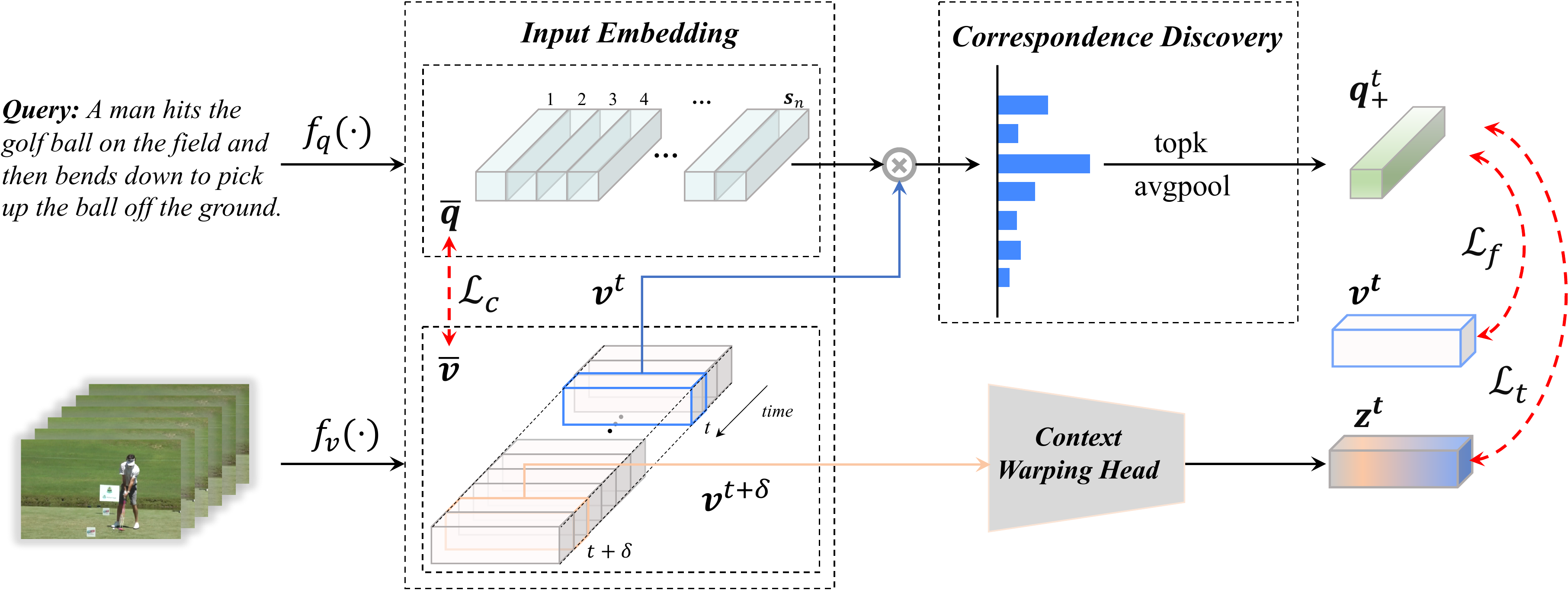}
	\vspace{-2mm}
	\caption{\footnotesize{An overview of LocVTP. $f_{v}(\cdot)$ and $f_{q}(\cdot)$ are video and language encoders, respectively. \textbf{1)} Coarse-grained contrastive loss $\mathcal{L}_{c}$ matches the global video and sentence representations $\overline{\boldsymbol{v}}$ and $\overline{\boldsymbol{q}}$. \textbf{2)} The clip-word correspondence is firstly built by similarity computing and then fine-grained contrastive loss $\mathcal{L}_{f}$ conducts detailed cross-modal alignment. Note that for clarity, we only present the correspondence discovery for the clip $\boldsymbol{v}^{t}$. \textbf{3)} A context warping head is employed to warp the contextual feature $\boldsymbol{v}^{t+\delta}$ and a temporal aware contrastive loss $\mathcal{L}_{t}$ is applied based on the warped feature $\boldsymbol{z}^{t}$.}}
	\label{fig:pipeline}
	\vspace{-2mm}
\end{figure}
\subsection{Overview of LocVTP}\label{sec:3.1}

An overview of LocVTP is illustrated in Fig.~\ref{fig:pipeline}. We firstly feed the video and language modalities to their respective encoders $f_{v}(\cdot)$ and $f_{q}(\cdot)$ to obtain embedded features. We follow the sparse sampling spirit in \cite{lei2021less} and sample $T$ clips for each video, yielding the encoded video $\boldsymbol{v}= \{\boldsymbol{v}^{t}\}_{t=1}^{T}$, where $\boldsymbol{v}^{t}\in \mathbb{R}^{D}$ is the $t^{th}$ clip feature and $D$ is the feature dimension. The text embedding is represented as $\boldsymbol{q}= \{\boldsymbol{q}^{s}\}_{s=1}^{S_{q}}$, where $\boldsymbol{q}^{s}\in \mathbb{R}^{D}$ is the $s^{th}$ word embedding and $S_q$ is the word length of $\boldsymbol{q}$.

Three types of contrastive methods are then performed to learn cross-modal features: 1) The coarse-grained contrastive loss builds the video-sentence level alignment; 2) A correspondence discovery strategy is proposed to build clip-word relations, based on which the fine-grained contrastive loss is applied; 3) Temporal aware contrastive loss with the context warping pretext task is proposed to encode temporal information into video representations.

\subsection{Coarse-grained Contrastive Learning}\label{sec:3.2}
We firstly conduct contrastive alignment at the global video-sentence level. Specifically, to obtain the video and sentence level features, we average pool $\boldsymbol{v}$ and $\boldsymbol{q}$ along the temporal and word index dimension, respectively. The global features are represented as $\overline{\boldsymbol{v}}$, $\overline{\boldsymbol{q}}\in \mathbb{R}^{D}$. Then we formulate this video-sentence alignment into the contrastive framework~\cite{he2020momentum} as follows:
\begin{equation}
\mathcal{L}_{c}=-\log \frac{\exp \left(\overline{\boldsymbol{v}} {\cdot} \overline{\boldsymbol{q}} / \tau\right)}{\sum_{i=1}^{N} \exp \left(\overline{\boldsymbol{v}} {\cdot} \overline{\boldsymbol{q}}_{i}  / \tau\right)}, \label{equ:1}
\end{equation}
\noindent where $\overline{\boldsymbol{q}}_{i}, i \in [1, N]$, is the sentence feature for other samples within the batch. $N$ denotes the batch size and $\tau$ is the temperature parameter. The coarse-grained contrastive loss $\mathcal{L}_{c}$ serves as a base loss to conduct video-sentence level constraint and induces a basic latent space where the detailed cross-modal matching is achieved. Though usually coarse and noisy, this latent space encodes prior for fine-grained clip-word correspondence discovery. In Section~\ref{subsec:4_6_1}, we design and analyze three potential ways to use this cross-modal matching prior.

\subsection{Fine-grained Contrastive Learning}\label{sec:3.3}

Beyond the coarse-grained video-sentence alignment, we propose to conduct contrastive learning in a fine-grained manner, \emph{i.e.}, clip-word matching. We contend that introducing such alignment learning into the pre-training stage could narrow down its gap with downstream localization tasks and calibrate the pre-trained feature to be more temporally aware.

\noindent \textbf{Clip-word correspondence discovery.} Before performing fine-grained contrastive learning, we firstly need to estimate the clip-word correspondences from video-sentence pairs. Thanks to the priors well established by the coarse-grained contrastive learning, we compute the cosine similarities between the video clips and their corresponding caption words in the pre-built latent space and choose the most similar $K$ words as the correspondence for each video clip. Note that we select multiple positive words rather than simply pick one with the highest similarity because individual words may have vague meanings while sense-group\footnote{A group or sequence of words conveying a particular meaning or idea in linguistics.\label{sensegroup}} conveys more precise information (cf. Section~\ref{subsec:4_6_2}).

Given the video sentence pair $\left\{\boldsymbol{v}, \boldsymbol{q}\right\}$, for the encoded $t^{th}$ video clip $\boldsymbol{v}^{t}$, we compute its cosine similarities with the $s^{th}$ word embedding $\boldsymbol{q}^{s}$ and apply the $\operatorname{topk}$ operation to select the most matched $K$ ones. Following~\cite{wang2021dense}, these $K$ selected items are average pooled to form the final positive sample:
\begin{equation}
\boldsymbol{q}_{+}^{t}=\operatorname{avgpool} \Big(\underset{s\in [1, S_{q}]}{\arg \operatorname{topk} } \left(\boldsymbol{v}^{t} {\cdot} \boldsymbol{q}^{s}\right)\Big), \label{equ:2}
\end{equation}
\noindent where $\boldsymbol{q}_{+}^{t}$ is the final positive sample for $\boldsymbol{v}^{t}$. $(\boldsymbol{u} {\cdot} \boldsymbol{v})=\boldsymbol{u}^{\top} \boldsymbol{v} /\|\boldsymbol{u}\|\|\boldsymbol{v}\|$ represents the cosine similarity between $\ell_{2}$ normalized $\boldsymbol{u}$ and $\boldsymbol{v}$. This process can be efficiently performed for all the video clips using matrix operations. 

\noindent \textbf{Fine-grained contrastive loss.} With the selected clip-word correspondence as positive pairs, we perform fine-grained representation learning following the cross-modal InfoNCE~\cite{he2020momentum} loss (cf. Figure~\ref{fig:posLossA}). The negative samples are taken from the other words within the batch. Therefore, the fine-grained contrastive loss is defined as follows.
\begin{equation}
\mathcal{L}_{f}=\frac{1}{T}\sum_{t=1}^{T}-\log \frac{ \exp (\boldsymbol{v}^{t} {\cdot} \boldsymbol{q}_{+}^{t} / \tau)}{\sum_{i=1}^{N} \sum_{s=1}^{S_{q_{i}}} \exp \left(\boldsymbol{v}^{t} {\cdot} \boldsymbol{q}_{i}^{s} / \tau\right)}, \label{equ:3}
\end{equation}

\noindent where $\boldsymbol{q}_{i}^{s}$ is the $s^{th}$ word feature of the $i^{th}$ sentence $\boldsymbol{q}_{i}$.

\subsection{Temporal aware Contrastive Learning}\label{sec:3.4}

\begin{figure}[t]
     \centering
     \begin{subfigure}[b]{0.4\textwidth}
         \centering
         \includegraphics[width=\textwidth]{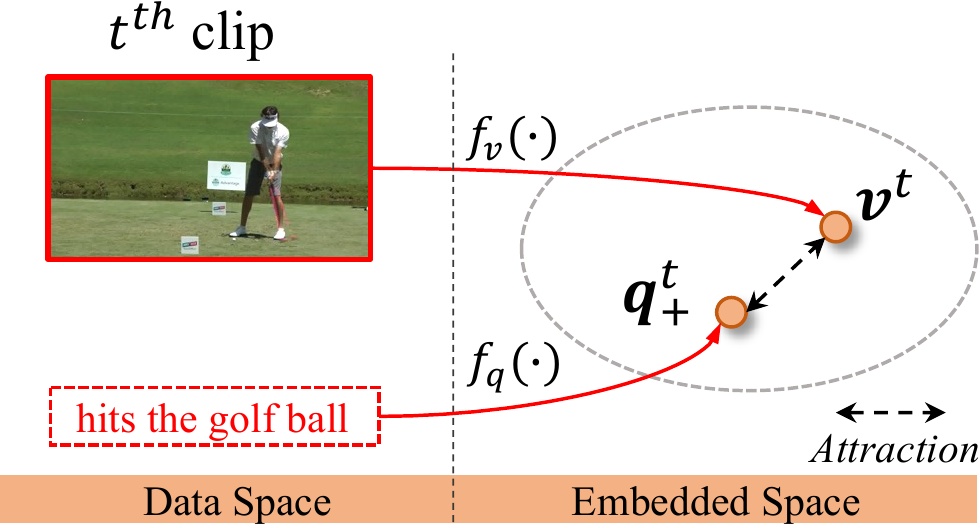}
         \caption{}
         \label{fig:posLossA}
     \end{subfigure}
     \hspace{1em}
     \begin{subfigure}[b]{0.4\textwidth}
         \centering
         \includegraphics[width=\textwidth]{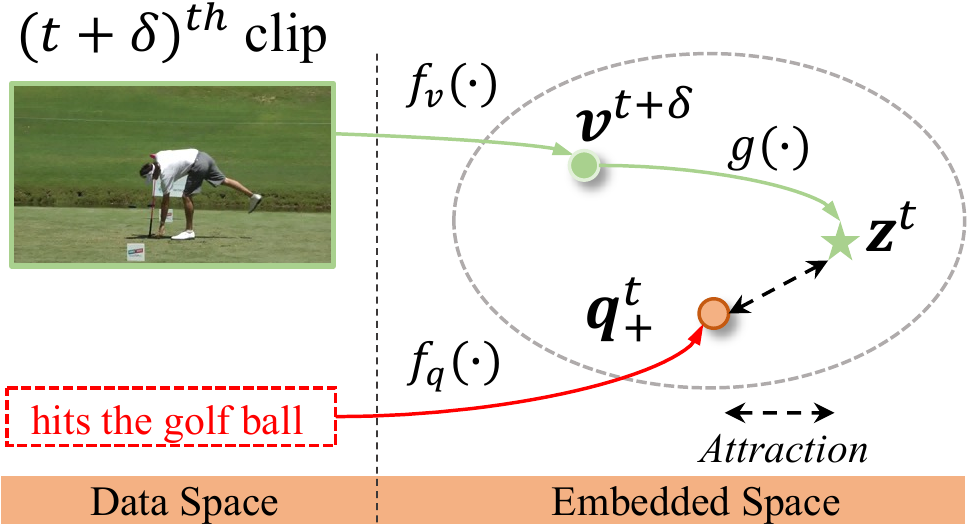}
         \caption{}
         \label{fig:posLossB}
     \end{subfigure}
    \vspace{-2mm}
    \caption{\footnotesize{Illustrations of (a) fine-grained contrastive Loss and (b) temporal aware contrastive loss. $\boldsymbol{v}^{t}$ is the $t^{th}$ clip of video $\boldsymbol{v}$. $\boldsymbol{q}_{+}^{t}$ is the pooled positive word features. $\boldsymbol{z}^{t}$ is the warped feature. We only present positive samples and omit negative ones.}}
    \vspace{-2mm}
    \label{fig:posLoss}
\end{figure}


Compared with the video-level retrieval task, which favors temporal invariant features~\cite{pan2021videomoco,qian2021spatiotemporal}, the clip-level localization task~\cite{chen2020rethinking,lu2019debug,xiao2021boundary,cao2021pursuit,xiao2021natural,yuan2021closer,zhang2021cola,cao2021deep,zhang2021synergic} prefers temporal aware video embeddings. Specifically, correlated actions in the same video should perceive each other. This characteristic is however not embodied in the aforementioned contrastive learning. 

\noindent \textbf{Context warping head.} To alleviate this, we set up a \emph{context-warping} operation to enforce the video clip to perceive the context. For the video clip $\boldsymbol{v}^{t}$ in a matched clip-word pair $\{\boldsymbol{v}^{t}, \boldsymbol{q}^{t}_{+}\}$ (cf. Section \ref{sec:3.3}), we warp its contextual video clip with $\delta$ temporal distance, \ie, $\boldsymbol{v}^{t+\delta}$, to ``reconstruct" itself. To supervise this warping process, we set up a temporal aware contrastive loss to maintain the established correspondence. Specifically, we propose a \textit{context warping head} $g(\cdot)$ to instantiate this warping process, by taking the context clip feature $\boldsymbol{v}^{t+\delta}$ and temporal distance $\delta$ as input.
\begin{equation}
\begin{aligned}
\boldsymbol{z}^{t} &=g(\boldsymbol{v}^{t+\delta}; \delta) \\
&=\operatorname{ReLU}\left(W[\boldsymbol{v}^{t+\delta}, \operatorname{sgn}(\delta),|\delta|]\right), \label{equ:4}
\end{aligned}
\end{equation}
\noindent where $\boldsymbol{z}^{t}$ is the warped feature. $W \in \mathbb{R}^{(D+2) \times D}$ are the trainable weights. $\delta$ is randomly sampled within the range of $[-\delta_{max}, \delta_{max}]$. $\operatorname{sgn}(\cdot)$ is the sign function which returns $1$ for positive values and $-1$ for negative ones. Here $\operatorname{sgn}(\delta)$ and $\lvert\delta\rvert$ indicate the direction and distance of the temporal difference $\delta$, respectively.

\noindent \textbf{Temporal aware contrastive loss.} Through the context warping head, the warped feature $\boldsymbol{z}^{t}$ should mimic the reference feature $\boldsymbol{v}^{t}$. Since $\boldsymbol{v}^{t}$ has the clip-word alignment with $\boldsymbol{q}^{t}_{+}$, such correspondence should be preserved between the warped feature $\boldsymbol{z}^{t}$ and $\boldsymbol{q}^{t}_{+}$ (cf. Fig.~\ref{fig:posLossB}).
\begin{equation}
\mathcal{L}_{t}=\frac{1}{T}\sum_{t=1}^{T}-\log \frac{ \exp (\boldsymbol{z}^{t} {\cdot} \boldsymbol{q}_{+}^{t} / \tau)}{\sum_{i=1}^{N} \sum_{s=1}^{S_{q_{i}}} \exp \left(\boldsymbol{z}^{t} {\cdot} \boldsymbol{q}_{i}^{s} / \tau\right)}. \label{equ:5}
\end{equation}

This process enforces video features to learn the ability of temporally reasoning, thus leading to more localization-friendly video features.

Integrating the above constraints, our final loss function is as follows. 
\begin{equation}
\mathcal{L} =  \lambda_{c}\mathcal{L}_{c} + \lambda_{f}\mathcal{L}_{f} + \lambda_{t}\mathcal{L}_{t}, \label{equ:lossAll}
\end{equation}

\noindent where $\lambda_{c}$, $\lambda_{f}$, and $\lambda_{t}$ balance the focus on different constraints during training.
\section{Experiments}\label{expes}

\subsection{Settings of Pre-training}\label{sec:4_1}
\noindent \textbf{Datasets.} We pre-trained our model on three public datasets: \textbf{1)} HowTo100M~\cite{miech2019howto100m}. It consists of more than 1.2M videos accompanied with ASR-generated speech transcription. The provided transcription is used to create video-sentence pairs separated by each timestamp. \textbf{2)} WebVid-2M~\cite{miech2019howto100m}. It contains about 2.5M well-aligned web video-text pairs. \textbf{3)} Google Conceptual Captions~\cite{sharma2018conceptual}. It contains 3.3M image and description pairs harvested from the web.

\noindent \textbf{Encoders.} Following \cite{bain2021frozen,wang2021object,yan2021video}, we adopted ViT-B/16~\cite{dosovitskiy2020image} with space-time attention~\cite{bertasius2021space} as the video encoder. The spatial attention weights in the transformer were initialized with ImageNet-21k pre-trained weights while the temporal attention weights were set to zero. We chose a lightweight DistilBERT~\cite{sanh2019distilbert} as the language encoder. Following~\cite{bain2021frozen,patrick2020support,tang2021decembert,amrani2020noise}, the language encoder was initialized with the weights pre-trained on English Wikipedia and Toronto Book Corpus.

\noindent \textbf{Implementation Details.} For the video in each video-sentence pair, we sampled 8 clips of 16 frames equidistantly and fed them to the video encoder to obtain clip-level features. All frames were resized to $224 \times 224$. For downstream transfer, we extracted video features with the well-trained model in a dense manner, \ie, every 16 consecutive frames were grouped to compute one clip feature.

Experiments were conducted on 64 V100 GPUs with a batch size of 256 and lasted for 200 epochs. We used Adam~\cite{loshchilov2017decoupled} with the initial learning rate $10^{-4}$ as the optimizer. The learning rate decayed by $0.1$ at the $100^{th}$ and $160^{th}$ epoch. Random flip, random crop, and color jitter for video data augmentation were included. The loss balance factors $\lambda_{c}$, $\lambda_{f}$, and $\lambda_{t}$ were set to 0.5, 1, 1, respectively. The temperature factor $\tau$ used in contrastive learning was set to 0.07 following \cite{wu2018unsupervised,radford2021learning} and $K$ in Eq.\eqref{equ:2} was set to 3. Features in all three contrastive losses were $\ell_{2}$-normalized before computation.
\subsection{Transfer Results on Video Retrieval} \label{sec:4_5}

\begin{table}[t]
	\footnotesize
	\centering
	\renewcommand\arraystretch{1.1}
	\setlength{\tabcolsep}{3pt}{
		\scalebox{0.8}{
		\begin{tabular}{r|lllllll}
			\multicolumn{1}{r|}{Method} & Vis Enc. Init.& Pre-trained Data &\#pairs & R@1 & R@5 & R@10 & MdR \\
			\shline
			UniVL~\cite{luo2020univl} & - & HowTo100M & 136M  &  21.2 & 49.6  &63.1 &6.0\\
			ClipBERT~\cite{lei2021less} & - & COCO, VGen & 5.6M & 22.0 & 46.8 & 59.9 & 6.0 \\
	        \textcolor{gray!80}{CE}~\cite{liu2019use} & \textcolor{gray!80}{Multi-modal} & \textcolor{gray!80}{HowTo100M} & \textcolor{gray!80}{136M} & \textcolor{gray!80}{20.9} & \textcolor{gray!80}{48.8} & \textcolor{gray!80}{62.4} & \textcolor{gray!80}{6.0}\\
			\textcolor{gray!80}{MMT}~\cite{gabeur2020multi} & \textcolor{gray!80}{Multi-modal} & \textcolor{gray!80}{HowTo100M} & \textcolor{gray!80}{136M} & \textcolor{gray!80}{26.6} & \textcolor{gray!80}{57.1} &  \textcolor{gray!80}{69.6} &  \textcolor{gray!80}{4.0} \\
			\textcolor{gray!80}{HIT}~\cite{liu2021hit} & \textcolor{gray!80}{Multi-modal} & \textcolor{gray!80}{HowTo100M} & \textcolor{gray!80}{136M} & \textcolor{gray!80}{30.7} & \textcolor{gray!80}{60.9} & \textcolor{gray!80}{73.2} & \textcolor{gray!80}{2.6} \\
		    Clip4clip$^\dag$~\cite{luo2021clip4clip} & CLIP &  HowTo100M & 136M & 44.5 & 71.4 & 81.6 & 2.0 \\
		    VideoClip~\cite{xu2021videoclip} & CLIP &  HowTo100M & 136M & 30.9 & 55.4 & 66.8 & -\\
		    OA-Trans~\cite{wang2021object} & CLIP & CC3M, WV2M & 5.5M & 40.9 & 70.4 & 80.3 & 2.0 \\
		    Frozen~\cite{bain2021frozen} & ImageNet &  CC3M, WV2M & 5.5M &  31.0 &   59.5 &  70.5 &  3.0 \\
			ActBERT~\cite{zhu2020actbert} & VisGenome & HowTo100M & 136M &16.3 &42.8 &56.9 &10.0 \\
			SupportSet~\cite{patrick2020support} & IG65M, ImageNet & HowTo100M & 136M & 30.1 & 58.5 & 69.3 & 3.0 \\
			HERO~\cite{li2020hero} & ImageNet, Kinetics & HowTo100M & 136M  &16.8 &43.4& 57.7 & - \\
			AVLnet~\cite{rouditchenko2020avlnet} & ImageNet, Kinetics & HowTo100M & 136M & 27.1 & 55.6 & 66.6 & 4.0 \\
			NoiseEstimation~\cite{amrani2020noise} & ImageNet, Kinetics  &HowTo100M & 136M & 17.4 & 41.6 &  53.6 & 8.0  \\
		    DECEMBER~\cite{tang2021decembert} & ImageNet, Kinetics & HowTo100M & 136M & 30.7 & 60.9 & 73.2 & 2.6 \\
		    OA-Trans~\cite{wang2021object} & ImageNet & CC3M, WV2M & 5.5M & 35.8  & 63.4 & 76.5 & 3.0 \\
		    RegionLearner$^\dag$~\cite{yan2021video} & ImageNet & CC3M, WV2M & 5.5M & 36.3 & 63.9 & 72.5 & 3.0\\ 
			\textbf{\textcolor{red}{LocVTP (Ours)}} & \textbf{\textcolor{red}{ImageNet}}  & \textbf{\textcolor{red}{HowTo100M}} & \textbf{\textcolor{red}{136M}} & \textbf{\textcolor{red}{37.4}}  & \textbf{\textcolor{red}{66.6}}  & \textbf{\textcolor{red}{80.5}}  & \textbf{\textcolor{red}{3.0}}  \\
			\textbf{\textcolor{orange}{LocVTP (Ours)}} & \textbf{\textcolor{orange}{CLIP}}  & \textbf{\textcolor{orange}{HowTo100M}} & \textbf{\textcolor{orange}{136M}} & \textbf{\textcolor{orange}{46.3}}  & \textbf{\textcolor{orange}{72.8}}  & \textbf{\textcolor{orange}{82.0}}  & \textbf{\textcolor{orange}{2.0}}  \\
			\textbf{\textcolor{blue}{LocVTP (Ours)}} & \textbf{\textcolor{blue}{ImageNet}}   & \textbf{\textcolor{blue}{CC3M,WV2M}} & \textbf{\textcolor{blue}{5.5M}}  & \textbf{\textcolor{blue}{36.5}}  & \textbf{\textcolor{blue}{64.3}}  & \textbf{\textcolor{blue}{76.8}}  & \textbf{\textcolor{blue}{3.0}}  \\
			\hline
			\multicolumn{8}{c}{\emph{{Zero-shot}}} \\
            SupportSet~\cite{patrick2020support} & IG65M, ImageNet & HowTo100M & 136M & 8.7 & 23.0 & 31.1 & 31.0     \\
		    Frozen~\cite{bain2021frozen} & ImageNet &  CC3M, WV2M &  5.5M & 18.7 & 39.5 & 51.6 & 10.0 \\
		    OA-Trans~\cite{wang2021object} & ImageNet & CC3M, WV2M & 5.5M & 23.4 & 47.5 & 55.6 & 8.0 \\
		    OA-Trans~\cite{wang2021object} & CLIP & CC3M, WV2M & 5.5M & 31.4 & 55.3 & 64.8 & 4.0\\
			\textbf{\textcolor{red}{LocVTP (Ours)}} & \textbf{\textcolor{red}{ImageNet}}  & \textbf{\textcolor{red}{HowTo100M}} & \textbf{\textcolor{red}{136M}} & \textbf{\textcolor{red}{24.7}}  & \textbf{\textcolor{red}{48.9}}  & \textbf{\textcolor{red}{56.1}}  & \textbf{\textcolor{red}{8.0}}  \\
			\textbf{\textcolor{orange}{LocVTP (Ours)}} & \textbf{\textcolor{orange}{CLIP}}  & \textbf{\textcolor{orange}{HowTo100M}} & \textbf{\textcolor{orange}{136M}} & \textbf{\textcolor{orange}{32.7}}  & \textbf{\textcolor{orange}{55.7}}  & \textbf{\textcolor{orange}{64.9}}  & \textbf{\textcolor{orange}{4.0}}  \\
			\textbf{\textcolor{blue}{LocVTP (Ours)}} & \textbf{\textcolor{blue}{ImageNet}}   & \textbf{\textcolor{blue}{CC3M,WV2M}} & \textbf{\textcolor{blue}{5.5M}}  & \textbf{\textcolor{blue}{22.1}}  & \textbf{\textcolor{blue}{48.0}}  & \textbf{\textcolor{blue}{55.3}}  & \textbf{\textcolor{blue}{8.0}}  \\
	\end{tabular}
	}}
	\caption{\footnotesize{\textbf{Video retrieval performance on MSR-VTT.} Vis Enc. Init.: Datasets used for pre-training visual encoders. Methods using multi-modal features are \textcolor{gray!80}{grayed out}. COCO: Coco Captions~\cite{chen2015microsoft}; VGen: Visual genome~\cite{krishna2017visual}; CC3M: Conceptual captions~\cite{sharma2018conceptual}; WV2M: WebVid-2M~\cite{bain2021frozen}; $^\dag$ denotes the technical report available on ArXiv.}}
	\label{tab:sotaVR}
	\vspace{-5mm}
\end{table}


\noindent \textbf{Datasets.} We evaluate our LocVTP on the widely-used benchmark \textbf{MSR-VTT} dataset~\cite{xu2016msr}. It is composed of 10K YouTube videos (9K for training and 1K for test). We report results on the train/test splits introduced in \cite{yu2018joint}. 

\noindent\textbf{Results.} \textbf{1)} As can be seen, we achieve state-of-the-art performance under both sets of data, \ie, HowTo100M and CC3M+WV2M. Specifically, when pre-trained on CC3M+WV2M, LocVTP outperforms Frozen~\cite{bain2021frozen} by an absolute lift of 4.8\% on R@5. \textbf{2)} It should be pointed out that although using RGB data only, our LocVTP achieves better performance than the methods using multi-modal expert features including motion, face, and speech, \eg, MMT~\cite{gabeur2020multi}. \textbf{3)} The recent work CLIP~\cite{radford2021learning} provides a stronger vision encoder and we also evaluate the performance based on it. It is shown that the CLIP's weights greatly improve the performance of LocVTP with R@5 achieving 72.8\%, surpassing top-performing CLIP-based methods. \textbf{4)} Our LocVTP also outperforms previous methods under the zero-shot setting, showing its generalization ability.
\begin{table*}[t]
	\footnotesize
	\centering
	\renewcommand\arraystretch{1.1}
	\resizebox{0.95\linewidth}{!}{
	\begin{tabular}{rc|cccc|cccc|cccc}
	\multirow{2}{*}{Models} & \multirow{2}{*}{PT Data} & \multicolumn{4}{c|}{ANet Captions} & \multicolumn{4}{c|}{Charades-STA} & \multicolumn{4}{c}{TACoS} \\
	~ & ~ & $R_1^{0.5}$ & $R_1^{0.7}$  & $R_5^{0.5}$ & $R_5^{0.7}$  & $R_1^{0.5}$ & $R_1^{0.7}$  & $R_5^{0.5}$ & $R_5^{0.7}$  & $R_1^{0.3}$ & $R_1^{0.5}$  & $R_5^{0.3}$ & $R_5^{0.5}$  \\
	\shline
	Sep.Pre.~\cite{zhang2020learning} & Kinetics & 44.4 & 27.1 & 77.6 & 62.1 & 39.7 & 23.8 & 79.6 & 52.3 & 37.2 & 25.6 & 58.2 & 45.5  \\ 
	\boldPurple{LocVTP (Ours)}   & \boldPurple{HT}$\ddag$ & \boldPurple{45.2} & \boldPurple{27.1} & \boldPurple{78.3} & \boldPurple{63.5} & \boldPurple{40.3} & \boldPurple{24.2} & \boldPurple{80.6} & \boldPurple{52.7} & \boldPurple{38.4}& \boldPurple{25.9} & \boldPurple{59.0} & \boldPurple{45.9} \\ 
	\hdashline
	VideoBERT$^\ast$\cite{sun2019videobert}  & HT & 37.2 & 21.0 & 66.7 & 53.6 & 32.7 & 19.5 & 68.1 & 46.2 & 33.8 & 22.2 & 51.6 & 41.0   \\
	MIL-NCE~\cite{miech2020end}  & HT & 41.8 & 24.5 & 73.5 & 57.7 & 37.0 & 21.2 & 74.3 & 50.4 & 35.1 & 23.5 & 53.7 & 42.5   \\
	UniVL~\cite{luo2020univl}  & HT & 42.2 & 25.4 & 75.3 & 60.5 & 38.2 & 22.7 & 77.2 & 51.4 & 35.7 & 23.7 & 55.8 & 43.7  \\
	SupportSet$^\ast$~\cite{patrick2020support} & HT & 41.9 & 25.2 & 74.7 & 58.3 & 37.4 & 21.6 & 75.6 & 50.9 & 35.5 & 23.5 & 54.2 & 43.2 \\
	\boldOrange{LocVTP (Ours)}  & \boldOrange{HT} & \boldOrange{48.2} & \boldOrange{30.5} & \boldOrange{80.1} & \boldOrange{64.7} & \boldOrange{43.6} & \boldOrange{26.3} & \boldOrange{81.9} & \boldOrange{55.3} & \boldOrange{41.6} & \boldOrange{28.9} & \boldOrange{61.4} & \boldOrange{47.6}   \\
	\hdashline
	Frozen~\cite{bain2021frozen}  & CC,WV & 43.3 & 25.8 & 75.8 & 59.3 & 38.8 & 22.9 & 77.6 & 50.3 & 35.7 & 23.5 & 54.4 & 43.7 \\
	OA-Trans$^\ast$\cite{wang2021object} & CC, WV & 43.6 & 25.9 & 76.5 & 60.2 & 39.2 & 22.6 & 78.5 & 50.8 & 35.2 & 22.5 & 53.4 & 42.6 \\
	\boldBlue{LocVTP (Ours)}  & \boldBlue{CC,WV}  & \boldBlue{46.1} & \boldBlue{27.6} & \boldBlue{78.9} & \boldBlue{63.7} & \boldBlue{41.2} & \boldBlue{24.8} & \boldBlue{81.3} & \boldBlue{53.5} & \boldBlue{39.6} & \boldBlue{27.8} & \boldBlue{60.4}  & \boldBlue{47.9} \\
	\hdashline
	December~\cite{tang2021decembert}  & HT & 43.0 & 25.1 & 76.0 & 60.2 & 37.2 & 21.6 & 78.3 & 50.6 & 34.8 & 22.9 & 55.1 & 43.9 \\
	ClipBERT~\cite{lei2021less}  & CO,VG & 42.6 & 24.6 & 75.3 & 59.7 & 37.0 & 20.8 & 77.7 & 50.2 & 33.7 & 21.0 & 54.3 & 43.3 \\
	\end{tabular}
	}
	\caption{\footnotesize{\textbf{Temporal grounding performances using pre-trained representations.} \texttt{Sep.Pre.}: separately pre-training, \ie, the video encoder supervisedly pre-trained on Kinetics and text encoder taken from BERT. We retrain the the temporal grounding method 2D-TAN~\cite{zhang2020learning} using the pre-trained features. HT: HowTo100M; CO: Coco Captions~\cite{chen2015microsoft}; VG: Visual genome~\cite{krishna2017visual}; CC: Conceptual captions~\cite{sharma2018conceptual}; WV: WebVid-2M~\cite{bain2021frozen}; \boldPurple{HT}$\ddag$: the subset of HowTo100M with the same training volume as Kinetics (300K pairs). Methods with $^\ast$ are not open source and we implement them ourselves. $^\dag$ denotes the technical report available on ArXiv.}}
	\vspace{-5mm}
	\label{tab:sotaVG}
\end{table*}
\subsection{Transfer Results on Temporal Grounding} \label{sec:4_2}
\noindent \textbf{Settings.} We validate the performance of pre-trained representations on temporal grounding, which aims to localize actions corresponding to the sentence from an untrimmed video. Specifically, we re-train the mainstream temporal grounding method 2D-TAN~\cite{zhang2020learning}\footnote{We choose 2D-TAN since it is relatively simple without too many dataset-specific parameters, which can fairly verify the effectiveness of pre-training features. Results on more advanced baselines are available in the supplementary material.} by only replacing the original input features with pre-trained ones. For ease of feature extraction, we choose representative VTP methods with publicly-available codes for comparisons.

\noindent \textbf{Datasets and Metrics.} \textbf{1)} ActivityNet Captions (ANet)~\cite{krishna2017dense}. It contains 20K untrimmed videos with 100K descriptions. By convention, we use 37,417 video-query pairs for training, 17,505 pairs for validation, and 17,031 pairs for testing. \textbf{2)} Charades-STA~\cite{gao2017tall}. Following the official split, 12,408 video-query pairs are used for training, and 3,720 pairs for testing. \textbf{3)} TACoS~\cite{regneri2013grounding}. It has 10,146 video-query pairs for training, 4,589 pairs for validation, and 4,083 pairs for testing.

Following prior works, we adopt “R@n, IoU@m” (abbreviated as $R^m_n$) as the metric, Specifically, $R^m_n$ is defined as the percentage of at least one of top-n retrieved moments having IoU with the ground-truth moment larger than $m$.

\noindent\textbf{Results.} \textbf{1)} As shown in Table~\ref{tab:sotaVG}, even trained with a much larger dataset, the current popular video-text pre-training frameworks achieve inferior performance compared to the separately pre-trained one. For example, Frozen~\cite{bain2021frozen} reaches 43.3\% at $R_1^{0.5}$ on ANet Captions, which is 1.1\% absolute value lower than the separately pre-trained counterpart. \textbf{2)} Either pre-trained on HowTo100M or CC + WV, our LocVTP outperforms both video-text pre-training methods by a large margin on all three datasets. For example, pre-trained on HowTo100M, LocVTP surpasses the separately pre-trained method by 3.8\% on $R_1^{0.5}$ of ANet Captions. \textbf{3)} For more fair comparisons, we sample a subset of HowTo100M by selecting the same training sample as Kinetics~\cite{kay2017kinetics} (300K training pairs), denoted as HT$^\ddag$ in Table~\ref{tab:sotaVG}. Although using noisy ASR captions, the results demonstrates that under the same training data volume, our LocVTP still shows better performance compared to the separately pre-trained method. This manifests that our performance improvement is brought by the sound architecture design rather than just the use of the large-scale dataset.

\subsection{Transfer Results on Action Step Localization} \label{sec:4_3}
\noindent \textbf{Settings.} In action step localization, each video belongs to a task and is annotated with multiple action steps described with short natural languages. The goal is to align each frame with the correct step in the text form. Following \cite{miech2019howto100m,zhu2020actbert,luo2020univl,yang2021taco}, we take \cite{zhukov2019cross} as the downstream localization method. Specifically, we compute the similarity between each frame and the action step descriptions in feature space to find the optimal frame-wise order of action steps for a video.

\begin{wraptable}{r}{0.4\textwidth}
    \centering  
		\footnotesize
		\renewcommand\arraystretch{1.1}
		\vspace{-6mm}
		\scalebox{0.9}{
		\begin{tabular}{r|x{25}x{25}}
			\multicolumn{1}{r|}{Method} & CTR & FA   \\
			\shline
			Zhukov \emph{et al.}~\cite{zhukov2019cross}  & 31.6 & - \\
	    	NN-Viterbi~\cite{alayrac2016unsupervised} & -- & 21.2 \\ 
	    	CBT~\cite{miech2020end} & -- & 53.9 \\
			MIL-NCE~\cite{miech2020end} & 40.5 & 61.0 \\
			ActBERT~\cite{zhu2020actbert} & 41.4 & 57.0 \\
			UniVL~\cite{luo2020univl}  & 42.0 & 70.0 \\
			TACo~\cite{yang2021taco} & 42.5 & 68.4 \\
			VideoClip~\cite{xu2021videoclip} & 47.3  & 68.7\\
			VLM~\cite{xu2021vlm} &  46.5  & 68.4\\
			\textbf{LocVTP (Ours)} & \textbf{51.7} &  \textbf{72.9}\\
		\end{tabular}}
		\label{tab:sotaStepLoc}
	\vspace{-3mm}
	\caption{\footnotesize{Comparison results of action step localization (CTR: average recall) and action segmentation (FA: frame-wise accuracy).}}%
	\label{tab:sotaStepLocActSeg} 
	\vspace{-6mm}
\end{wraptable}

\noindent \textbf{Datasets and Metrics.} We experiment on the instructional video dataset CrossTask~\cite{zhukov2019cross}, which includes 83 tasks and 4.7K videos. Each task is described with an ordered list of steps with manual natural language descriptions. We perform the same evaluation protocol as in \cite{zhukov2019cross} by reporting the average recall (CTR).

\noindent\textbf{Results.} Table~\ref{tab:sotaStepLocActSeg} reports the action step localization performance on CrossTask dataset. Our LocVTP pre-trained feature achieves state-of-the-art performance with CTR reaching 51.7\%, surpassing the previous method VideoClip by 4.4\%. Our competitive performance demonstrates that LocVTP features can effectively perceive detailed action steps.
\subsection{Transfer Results on Action Segmentation} \label{sec:4_4}

\noindent\textbf{Settings.} We assess our LocVTP on action segmentation, which aims to predict the action label frame-wisely for each video frame. It is a pure vision task without the use of the text encoder.  Following \cite{yang2021taco,luo2020univl,zhu2020actbert}, we encode the input video frames with the well-trained video encoder and apply a linear classifier upon the features to predict action labels. 

\noindent \textbf{Datasets and Metrics.} We conduct experiments on the widely used COIN dataset~\cite{tang2019coin} and the frame-wise accuracy (FA) is taken as the evaluation metric.

\noindent\textbf{Results.} As shown in Table~\ref{tab:sotaStepLocActSeg}, our LocVTP achieves state-of-the-art performance with FA reaching 72.9\%. This further demonstrates the superiority of our feature in localization tasks even in the absence of language guidance.

\subsection{Ablation Study on Training Objective\protect\footnote{If not specified, all ablation studies are conducted on the downstream temporal grounding task at ActivityNet Captions dataset. We use LocVTP pre-trained on HowTo100M with ImageNet initialization.\label{ablations}}}\label{subsec:4_6_1}

\noindent \textbf{Training Strategy}. Coarse-grained contrastive alignment loss $\mathcal{L}_{c}$ provides a basic cross-modal matching prior and we introduce three potential ways to use it: \textbf{1)} \textit{multi-stage training}: first perform coarse-grained training and then use the trained model to initialize other stages. \textbf{2)} \textit{warm-up training}: decrease $\lambda_{c}$ exponentially from 1 to 0 throughout the training process. \textbf{3)} \textit{weighted training}: set $\lambda_{c}$ to a constant value. Here we set $\lambda_{c} = 0.5$. As shown in Table~\ref{tab:trainStrategy}, we find the weighted training strategy achieves the best performance and warm-up training is slightly behind. Multi-stage training is the least effective one. 

\noindent \textbf{Loss Component}. We present the loss component ablations in Table~\ref{tab:lossComp}. As shown, both fine-grained loss $\mathcal{L}_{f}$ and temporal aware loss $\mathcal{L}_{t}$ are crucial. For example, compared to the full version (exp.\#1), removing $\mathcal{L}_{f}$ and $\mathcal{L}_{t}$ brings about 1.4\% and 1.5\% performance degradation on the $R_1^{0.5}$ metric, respectively.

\noindent \textbf{More downstream temporal grounding baselines.} We take another temporal grounding method CSMGAN~\cite{liu2020jointly} as the downstream baseline. As shown in Table.~\ref{tab:baseline}, our LocVTP pre-trained feature consistently benefits this more advanced baseline.

\begin{table}[t]
\footnotesize
\centering
	\begin{subtable}[h]{0.3\textwidth}
		\centering
		\footnotesize
		\renewcommand\arraystretch{1.1}
        \scalebox{0.9}{
		\begin{tabular}{x{45}|x{28}x{28}}
		    \footnotesize
			Mode &   $R^{0.5}_1$ & $R^{0.7}_1$ \\
			\shline
			\emph{multi-stage}  & 47.4 & 29.7  \\
			\emph{warm-up}      & 47.7 & 30.1  \\
			\emph{weighted}  & \textbf{48.2}& \textbf{30.5} \\ 
			\multicolumn{3}{c}{~}\\ 
		\end{tabular}
        }
	    \vspace{-2mm}
		\caption{}
		\vspace{-1mm}
		\label{tab:trainStrategy}
	\end{subtable}
	\begin{subtable}[h]{0.38\textwidth}
		\centering
		\footnotesize
		\renewcommand\arraystretch{1.1}
        \scalebox{0.9}{
		\begin{tabular}{x{12}|x{12}x{12}x{12}|x{35}x{35}}
		\footnotesize
	    ~& $\mathcal{L}_{c}$ & $\mathcal{L}_{f}$ & $\mathcal{L}_{t}$ & $R^{0.5}_1$ & $R^{0.7}_1$  \\
			\shline
	 \#1	&	\cmark & \cmark& \cmark & \textbf{48.2}& \textbf{30.5}  \\ 
	 \#2	&	\cmark & \cmark&  			& {46.7}$_{\color{mygray}{-1.5}}$ & {29.4}$_{\color{mygray}{-1.1}}$    \\
	 \#3	&	\cmark &    		& \cmark & {46.8}$_{\color{mygray}{-1.4}}$ & {29.6}$_{\color{mygray}{-0.9}}$  \\
	 \#4	&	\cmark &           &             & 45.6$_{\color{mygray}{-2.6}}$ & 29.0$_{\color{mygray}{-1.5}}$  \\
		\end{tabular}
        }
	    \vspace{-2mm}
		\caption{}
		\vspace{-1mm}
		\label{tab:lossComp}
	\end{subtable}
	\begin{subtable}[h]{0.3\textwidth}
		\centering
		\footnotesize
		\renewcommand\arraystretch{1.1}
        \scalebox{0.9}{
		\begin{tabular}{r|x{25}x{25}}
		    \footnotesize
			Method & $R^{0.5}_1$ & $R^{0.7}_1$ \\
			\shline
			Sep.Pre.~\cite{liu2020jointly}  & 48.9 & 29.0  \\
			Frozen~\cite{bain2021frozen} & 47.3 & 26.8  \\
			LocVTP & \textbf{53.9} & \textbf{34.6} \\
			\multicolumn{3}{c}{~}\\ 
		\end{tabular}
        }
	    \vspace{-2mm}
		\caption{}
		\vspace{-1mm}
		\label{tab:baseline}
	\end{subtable}	
	\vspace{-2mm}
	\caption{\footnotesize{\textbf{Ablations studies} of (a) training strategies; (b) loss component; (c) comparison results on temporal grounding method CSMGAN~\cite{liu2020jointly}. \texttt{Sep.Pre.}: separately pre-training, \ie, the video encoder supervisedly pre-trained on Kinetics and text encoder taken from BERT.}}
	\vspace{-5mm}
\end{table}

\subsection{Ablations on Fine-grained Contrastive Loss}\label{subsec:4_6_2}

\noindent \textbf{Correspondence Discovery Strategies}. We experiment four potential strategies to extract cross-modal correspondences: \textbf{1)} \emph{random}: randomly select $K$ words for each clip; \textbf{2)} \emph{2d-topk}: select the most similar $K \times T$ clip-word pairs; \textbf{3)} \emph{word-topk}: select the most similar $K$ clips for each word; \textbf{4)} \emph{clip-topk}: select the most similar $K$ words for each clip, namely the method illustrated in Section~\ref{sec:3.3}. 
As indicated in Table.~\ref{tab:MatchStrategy}, the \textit{random} and \textit{2d-topk} matching strategies are the two worst options. For the \textit{word-topk} matching, it is also sub-optimal, which can be attributed to the possibility of introducing words without concrete meanings (\eg, articles or pronouns) into matched pairs.

\noindent \textbf{Number of Selected Pairs $K$}. We further ablate the hyper-parameter $K$ used in the \emph{clip-topk} strategy. Table~\ref{tab:AblaK} shows that the performance saturates at $K=3$ and slightly decreases for $K=4$. We conjecture that this may be because too few words have vague meanings while too large $K$ value leads to the inability to establish accurate correspondences.

\subsection{Ablations on Temporal aware Contrastive Loss}
\begin{table}[t]
\footnotesize
\centering
	\begin{subtable}[h]{0.32\textwidth}
		\centering
		\footnotesize
		\renewcommand\arraystretch{1.1}
		\scalebox{0.85}{
		\begin{tabular}{l|x{30}x{30}}
		    & $R_1^{0.5}$ & $R_1^{0.7}$  \\
			\shline
			\emph{random}  & 42.0 & 25.8  \\ 
		    \emph{2d-topk}  & 44.8  & 27.2   \\ 
			\emph{word-topk}& 47.0 & 28.7 \\ 
			\emph{clip-topk} & \textbf{48.2}& \textbf{30.5}  \\
		\end{tabular}
	}
	    \vspace{-2mm}
		\caption{}
		\vspace{-1mm}
		\label{tab:MatchStrategy}
	\end{subtable}
	\hfill
	\begin{subtable}[h]{0.25\textwidth}
		\centering
	\footnotesize
	\renewcommand\arraystretch{1.1}
	\scalebox{0.85}{	
	\begin{tabular}{x{10}|x{30}x{30}}
		\multicolumn{1}{l|}{$K$}& $R_1^{0.5}$ & $R_1^{0.7}$   \\
		\shline
		1  & 46.3 & 28.8  \\ 
		2  & 47.5 & 29.7    \\ 
		3  & \textbf{48.2}& \textbf{30.5}    \\ 
		4  & 48.0 & 29.8    \\ 
	\end{tabular}
	}
	\vspace{-2mm}
	\caption{}
	\vspace{-1mm}
	\label{tab:AblaK}
	\end{subtable}
	\hfill
	\begin{subtable}[h]{0.35\textwidth}
		\centering
		\footnotesize
		\renewcommand\arraystretch{1.1}
		\scalebox{0.85}{
		\begin{tabular}{x{20}x{17}|x{30}x{30}}
			$\operatorname{sgn}(\delta)$ & $\lvert\delta\rvert$  & $R_1^{0.5}$ & $R_1^{0.7}$   \\
			\shline
			\cmark& \cmark &\textbf{48.2}& \textbf{30.5}  \\ 
			\cmark& \xmark & 47.3  & 29.3   \\ 
			\xmark& \cmark & 47.1  & 29.0  \\ 
			\xmark& \xmark & 46.2  & 28.1 \\ 
		\end{tabular}}   
		\vspace{-2mm}
		\caption{}
		\vspace{-1mm}
		\label{tab:projAbla}
	\end{subtable}\hfill
	\begin{subtable}[h]{0.23\textwidth}
	\centering
	\footnotesize
	\renewcommand\arraystretch{1.1}
		\scalebox{0.85}{
		\begin{tabular}{x{15}|x{28}x{28}}
			  \multicolumn{1}{c|}{$\delta_{max}$} & $R_1^{0.5}$ & $R_1^{0.7}$ \\
			 \shline
			  2  & 47.6 & 29.2  \\ 
		      3 & 47.8 & 29.5 \\ 
		      4 & \textbf{48.2}  & \textbf{30.5}     \\ 
		      5 & 47.7  & 28.9    \\ 
	    \end{tabular}} 
	\vspace{-2mm}
	\caption{}
	\label{tab:shiftRatio}
	\end{subtable}\hfill
	\begin{subtable}[h]{0.3\textwidth}
	\centering
	\footnotesize
	\renewcommand\arraystretch{1.1}
		\scalebox{0.85}{
		\begin{tabular}{r|x{28}x{28}} 
		  \multicolumn{1}{c|}{$\mathcal{L}_{c}$ Mode} & $R_1^{0.5}$ & $R_1^{0.7}$  \\
			\shline
		   intra-modal & 47.7  & 29.8  \\ 
		   cross-modal & \textbf{48.2}& \textbf{30.5} \\ 
		   \multicolumn{3}{c}{~}\\ 
		   \multicolumn{3}{c}{~}\\ 
	    \end{tabular}
	}
	\vspace{-2mm}
	\caption{}
	\label{tab:intraVScross}
	\end{subtable}\hfill
	\begin{subtable}[h]{0.4\textwidth}
	\centering
	\footnotesize
	\renewcommand\arraystretch{1.1}
	\scalebox{0.85}{
		\begin{tabular}{l|x{25}x{25}x{25}x{25}}
			\multicolumn{1}{l|}{Method}& $Accu_o$ & $Accu_d$  \\
			\shline
			LocVTP (w/ $\mathcal{L}_{c}$)  & \textbf{72.8} & \textbf{58.2} \\ 
		    LocVTP (w/o $\mathcal{L}_{c}$)  & 69.0 & 56.5\\ 
		    UniVL~\cite{luo2020univl} & 64.2 & 52.8\\ 
		    MIL-NCE~\cite{miech2020end}	& 61.3 & 51.4\\ 
	\end{tabular}
	}
	\vspace{-2mm}
	\caption{}
	\label{tab:locEval}
	\end{subtable}
	\vspace{-2mm}
	\caption{\footnotesize{\textbf{Ablations studies} of (a) correspondence discovery strategies; (b) selected pair number $K$; (c) context projection head. $\operatorname{sgn}(\delta)$, $\lvert\delta\rvert$ denotes the direction and distance; (d) the maximum bias distance; (e) intra-modal \emph{v.s.} cross-modal $\mathcal{L}_{t}$; (f) linear localization accuracy. $Accu_o$ and $Accu_d$ are order and distance prediction accuracy.}}
	\vspace{-7mm}
\end{table}


\noindent \textbf{Context Projection Head Components.} In Eq.~\eqref{equ:4}, the warped feature is generated based on both the direction $\operatorname{sgn}(\delta)$ and distance $\lvert\delta\rvert$. Here we investigate eliminating either of them to see the difference. We observe in Table.~\ref{tab:projAbla} that removing either component decreases the performance, which indicates that both the direction and distance of bias $\delta$ are crucial for feature warping.

\noindent \textbf{Maximum Bias Distance $\delta_{max}$.} 
Here we ablate different values for $\delta_{max}$. From Table \ref{tab:shiftRatio}, we can see that $\delta_{max} = 4$ achieves the best performance. This may be because that small bias makes the model unable to perceive enough context, while a large bias makes contextual reasoning too difficult.

\noindent \textbf{Intra-modal \emph{v.s.} Cross-modal Constraint.} In Section.~\ref{sec:3.4}, given the matched clip-word pair $\{\boldsymbol{v}^{t}, \boldsymbol{q}^{t}_{+}\}$ and the warped feature $\boldsymbol{z}^{t}$, we force the \emph{cross-modal} supervision, \ie, $\boldsymbol{z}^{t} \leftrightarrow \boldsymbol{q}^{t}_{+}$. Here, we apply the temporal aware contrastive loss $\mathcal{L}_{t}$ in a \emph{intra-modal} manner which regards $\boldsymbol{z}^{t}$ and $\boldsymbol{v}^{t}$ as positive pairs, \ie, $\boldsymbol{z}^{t} \leftrightarrow \boldsymbol{v}^{t}$. The results in Table~\ref{tab:intraVScross} show that our adopted cross-modal mode outperforms the intra-modal one.

\noindent\textbf{Temporal Sensitivity Analysis.} As a sanity check, we devise two proxy tasks to evaluate the temporal sensitivity of pre-trained video features. As shown in Fig.~\ref{fig:loclizableEval}, $n$ equidistantly sampled clips from one video are fed into the frozen video backbone to extract their corresponding features. Two linear classifiers are trained to perform two tasks: \textit{order prediction} and \textit{distance estimation}. The first task predicts the temporal index while the second one estimates the temporal distance of two clips. The results in Table~\ref{tab:locEval} show that our LocVTP with temporal aware loss $\mathcal{L}_{t}$ outperforms the variant without it as well as two typical VTP methods (\ie, UniVL and MIL-NCE), which shows that $\mathcal{L}_{t}$ clearly contributes to the localization ability.
\vspace{-4mm}
\subsection{Visualization\protect \footnote{More visualizations are left in the supplementary materials.\label{supple}}} \label{sec:4_7}

\begin{figure}[t]
    \footnotesize
	\centering
	\begin{subfigure}[b]{0.23\textwidth}
		\centering
		\includegraphics[width=0.95\textwidth]{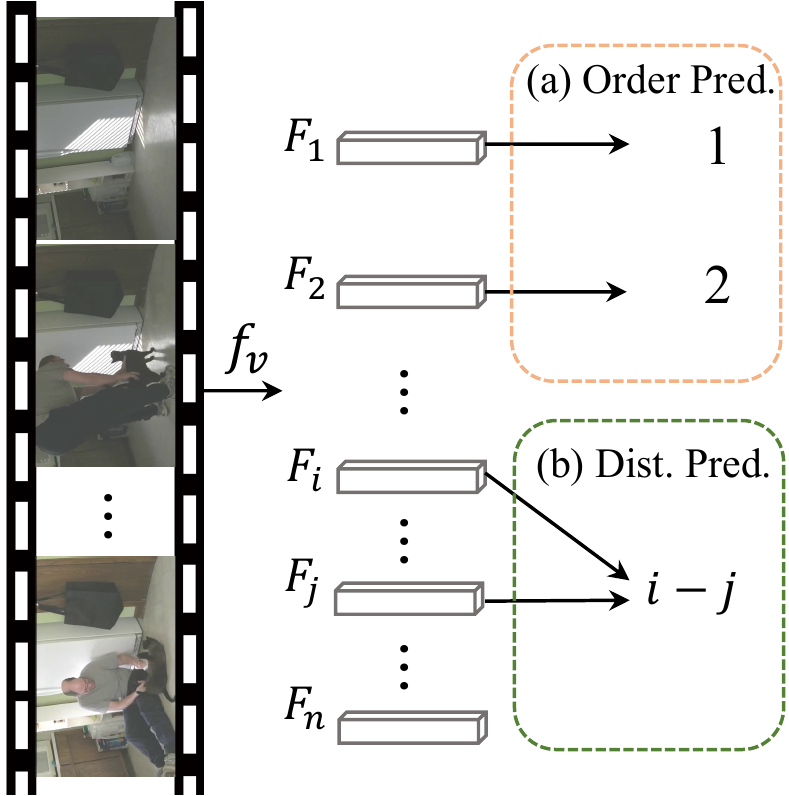}
		\caption{}
		\label{fig:loclizableEval}
	\end{subfigure}\hfill
	\begin{subfigure}[b]{0.47\textwidth}
		\centering		
        \includegraphics[width=0.82\textwidth]{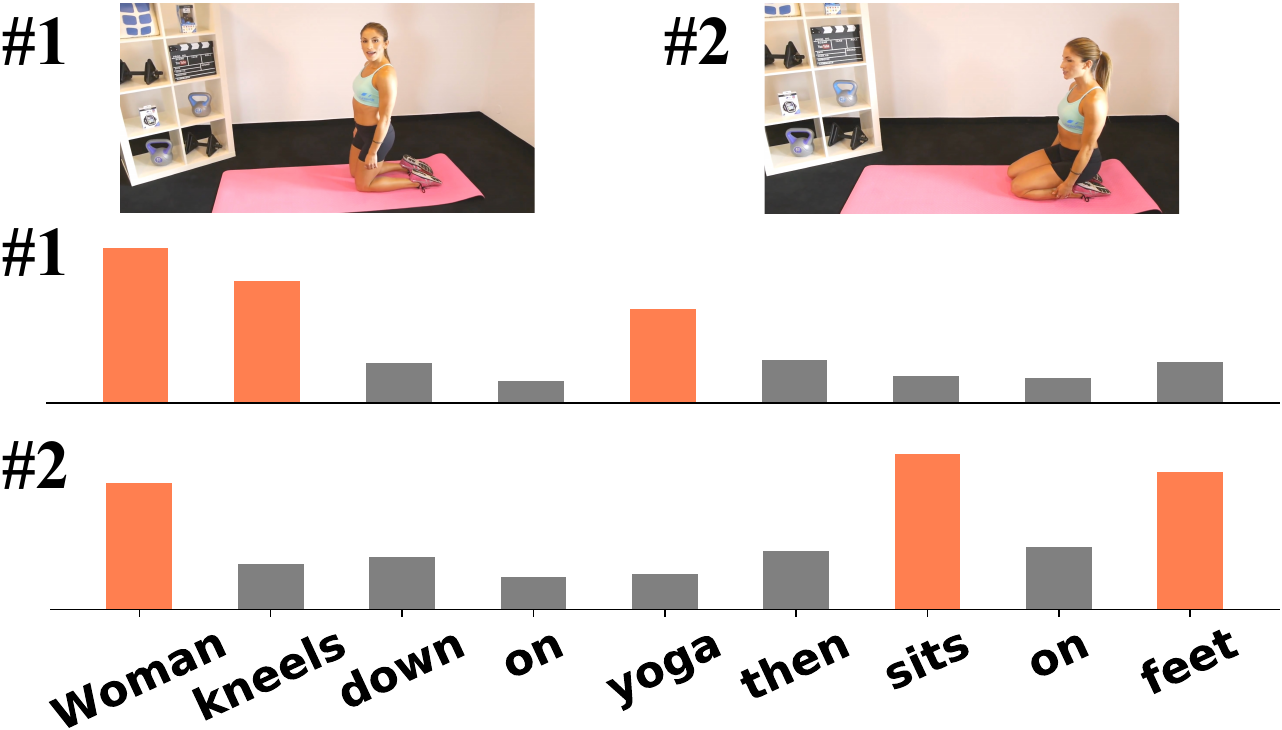}
		\caption{}
		\label{fig:fineMatch}
	\end{subfigure}\hfill
	\begin{subfigure}[b]{0.28\textwidth}
		\centering
		\includegraphics[width=0.95\textwidth]{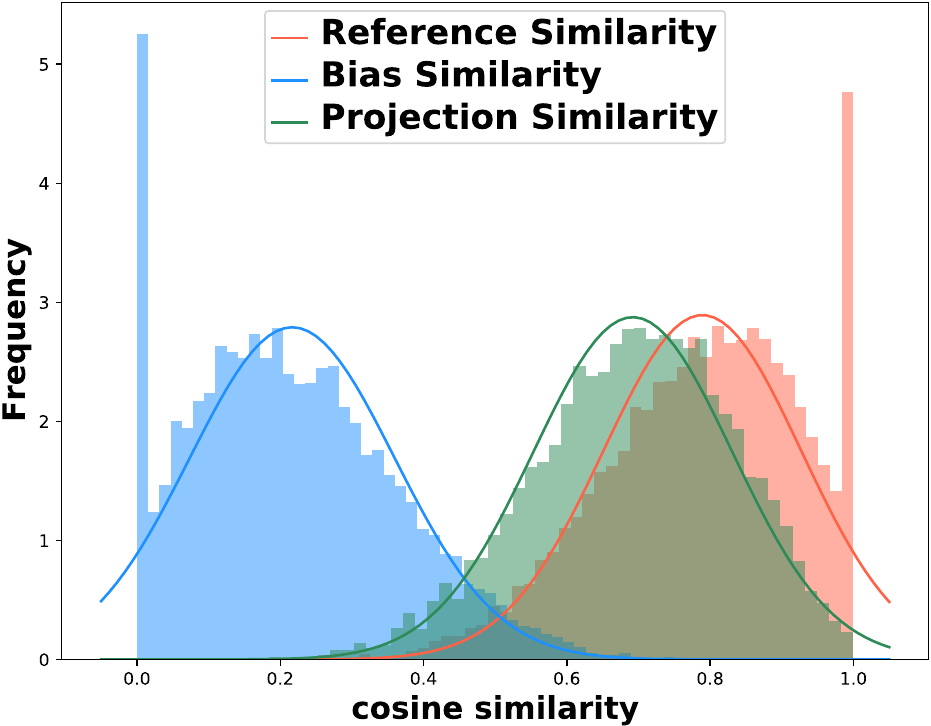}
		\caption{}
		\label{fig:distribute}
	\end{subfigure}
    \vspace{-2mm}
	\caption{\footnotesize{(a) Linear localization evaluations including order and distance prediction; (b) Cross-modal correspondence visualizations. Top \emph{K} responsive words are marked with \textcolor[RGB]{255,124,76}{\textbf{red}}. (c) Gaussian distributions of the \textcolor[RGB]{255,99,71}{\textbf{reference}}, \textcolor[RGB]{30,144,255}{\textbf{biased}}, and \textcolor[RGB]{46,139,87}{\textbf{projected}} similarities.}}
	\vspace{-1mm}
\end{figure}

\noindent\textbf{Cross-modal Correspondence Visualizations.} Fig.~\ref{fig:fineMatch} shows two frames\footnote{Here we use “frame” to indicate the center frame of a video snippet. \label{frame_path}} and their corresponding similarity scores with caption words. The top \emph{K} highest scored words are marked with red ($K=3$). 
Frame \#1 and frame \#2 have similar appearance views yet correspond to different action processes. Our method pinpoints the subtle differences and accurately finds the most relevant words.


\begin{figure}[t]
    \footnotesize
	\centering
	\begin{subfigure}[b]{0.22\textwidth}
		\centering
		\includegraphics[width=\textwidth]{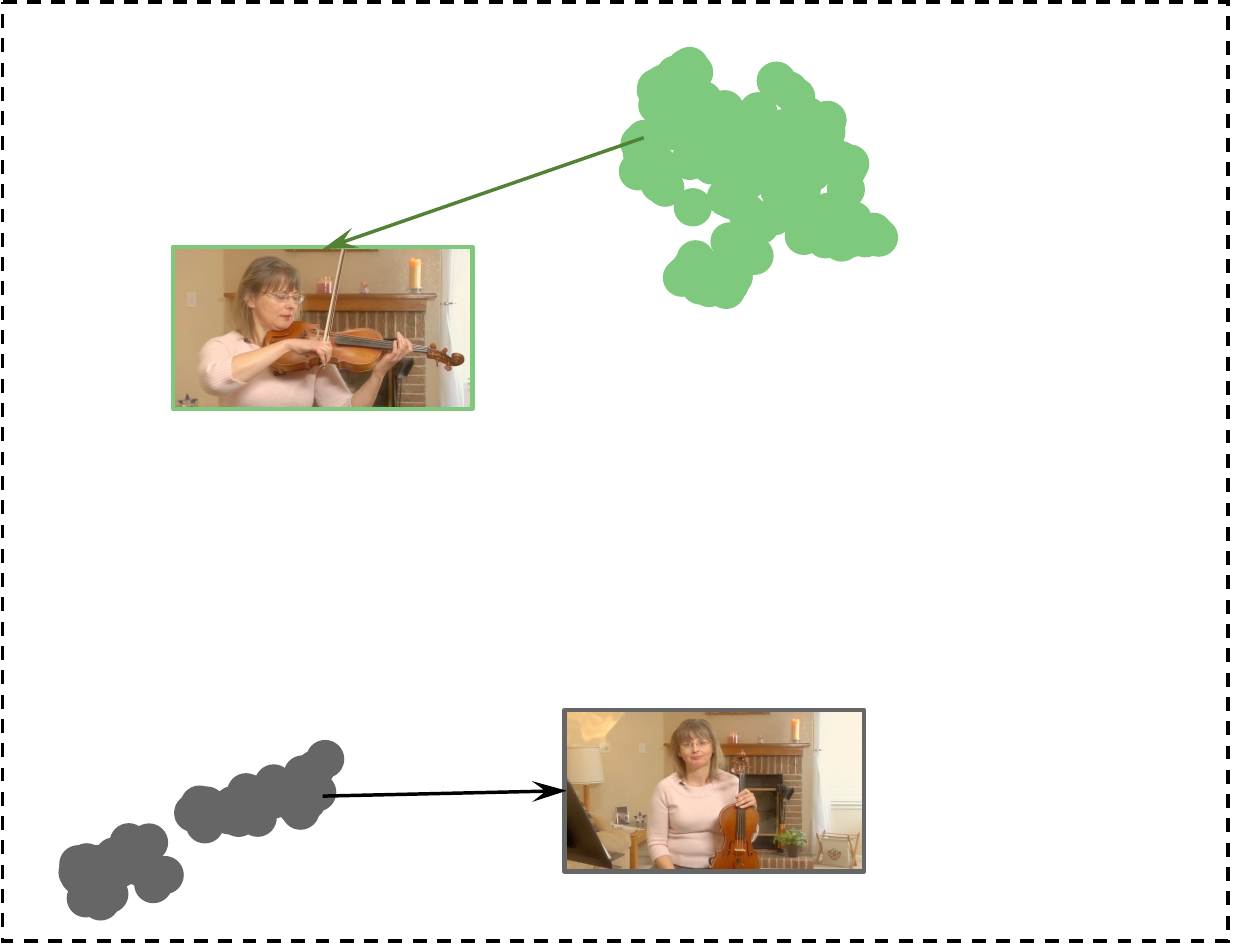}
		\caption{\footnotesize{Ours(\emph{w/}$\mathcal{L}_{t}$)}}
		\label{fig:umapa}
	\end{subfigure}
	\begin{subfigure}[b]{0.22\textwidth}
		\centering
		\includegraphics[width=\textwidth]{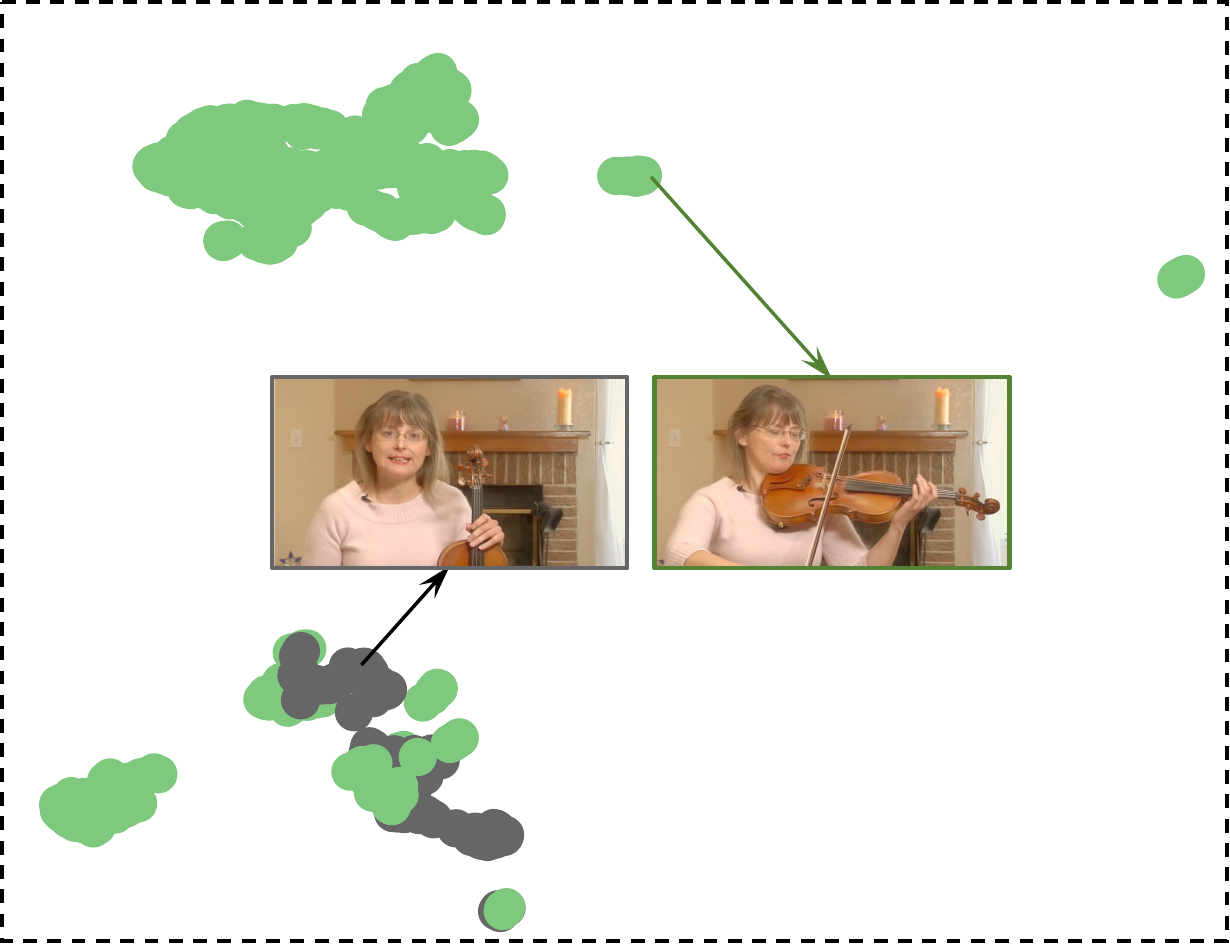}
		\caption{\footnotesize{Ours(\emph{w/o}$\mathcal{L}_{t}$)}}
		\label{fig:umapb}
	\end{subfigure}
	\begin{subfigure}[b]{0.22\textwidth} 
		\centering
		\includegraphics[width=\textwidth]{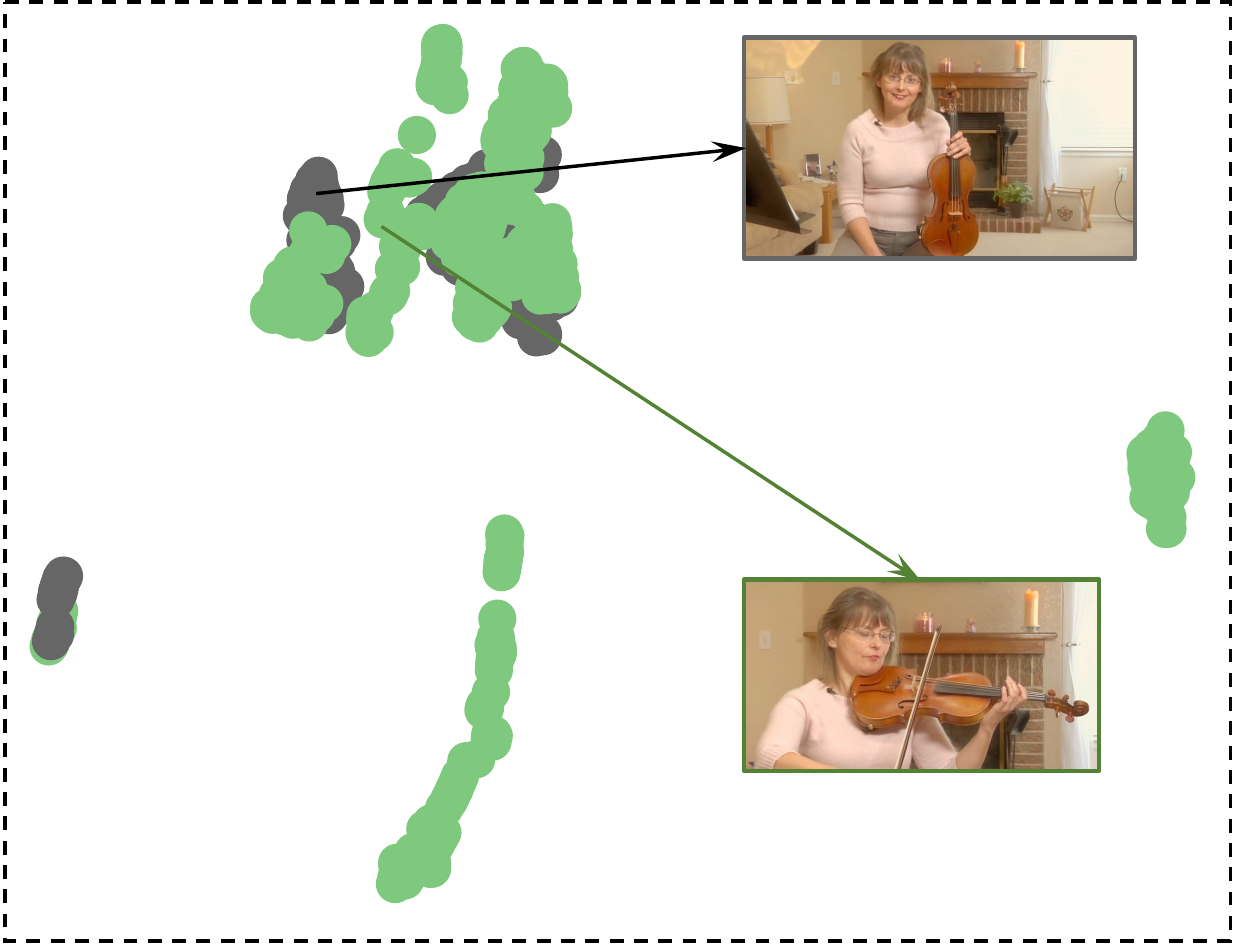}
		\caption{\footnotesize{UniVL}}
		\label{fig:umapc}
	\end{subfigure}
    \begin{subfigure}[b]{0.22\textwidth}
    	\centering
    	\includegraphics[width=\textwidth]{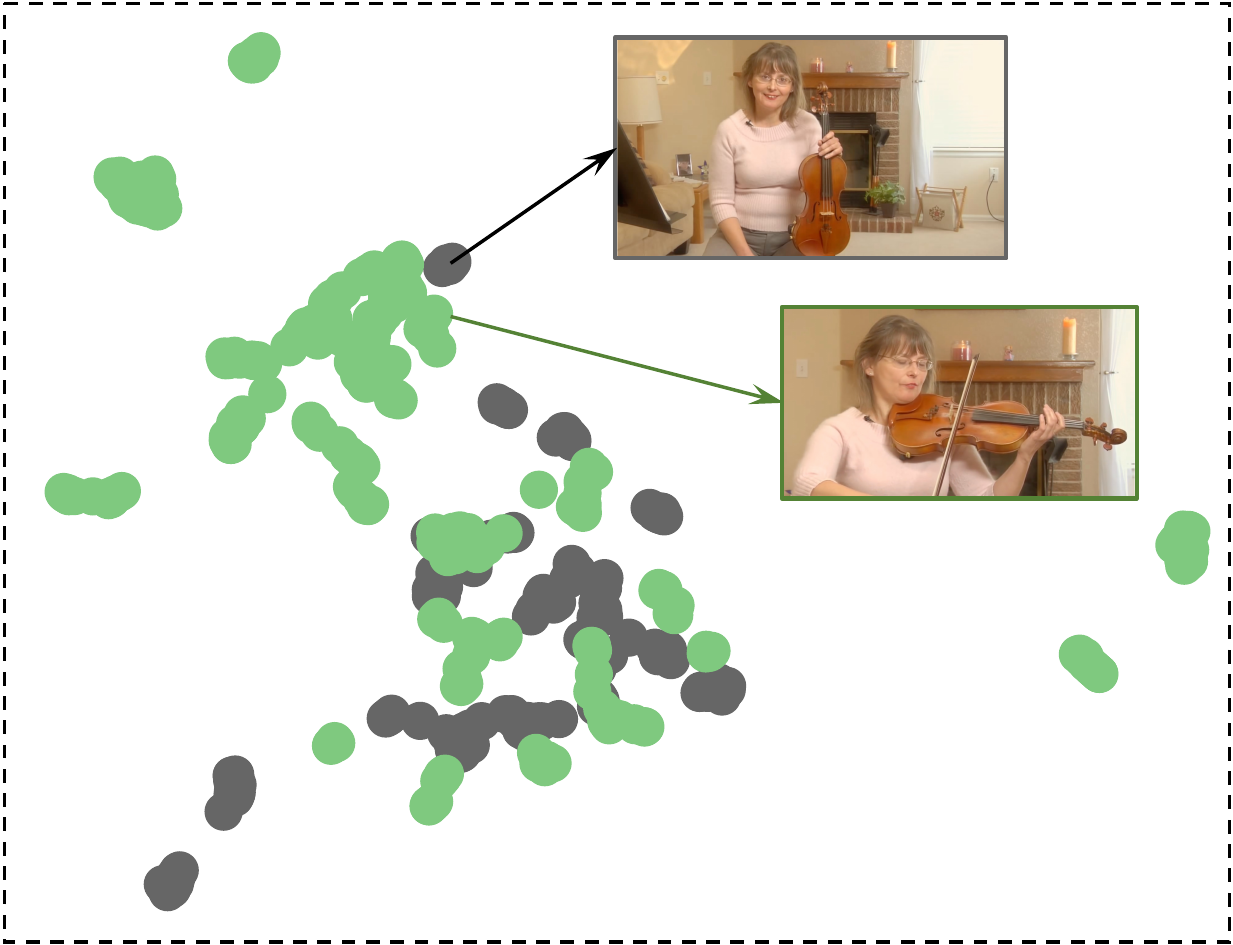}
    	\caption{\footnotesize{MIL-NCE}}
    	\label{fig:umapd}
    \end{subfigure}
    \vspace{-2mm}
	\caption{\footnotesize{UMAP visualizations. Clips corresponding to ground-truth caption are marked with \textcolor[RGB]{127,201,126}{\textbf{green}} while others are with \textcolor[RGB]{102,102,102}{\textbf{gray}}.}}
	\vspace{-4mm}
	\label{fig:umap}
\end{figure}
 

\noindent\textbf{UMAP Visualizations.} As shown in Fig.~\ref{fig:umap}, we provide UMAP~\cite{mcinnes2018umap} visualizations for \emph{fused} multi-modal features, which are generated by multiplying the extracted video feature by one query feature. With the temporal aware loss $\mathcal{L}_{t}$, our LocVTP shows more separable distributions compared with LocVTP \emph{w/o} $\mathcal{L}_{t}$, manifesting that $\mathcal{L}_{t}$ helps distinguish action-of-interest from background.

\noindent\textbf{Similarity Distribution Visualizations.} In Eq.\eqref{equ:4}, context projection head warps contextual clip $\boldsymbol{v}^{t+\delta}$ to the reference one $\boldsymbol{v}^{t}$. Here we collect 10K paired training samples and compute three sets of cosine similarities: reference similarity $\small{(\boldsymbol{v}^{t}, \boldsymbol{q}_{+}^{t})}$, bias similarity $\small{(\boldsymbol{v}^{t+\delta}, \boldsymbol{q}_{+}^{t})}$, and projection similarity $\small{(\boldsymbol{z}^{t}, \boldsymbol{q}_{+}^{t})}$. Fig.~\ref{fig:distribute} plots the histogram of these similarities. We can see that the distribution of projection similarity is close to that of reference similarity while far away from that of bias similarity. This demonstrates that our context projection head can effectively warp contextual features conditioned on the temporal information.


\section{Conclusions}
\vspace{-2mm}
In this paper, we propose LocVTP, the first video-text pre-training framework for temporal localization tasks. Specifically, we apply cross-modal contrastive learning at both coarse-grained video-sentence and fine-grained clip-word levels. Besides, we propose a context warping pretext task and a temporal aware contrastive loss to enhance the temporal awareness of video features. Experimental results show that LocVTP achieves state-of-the-art performance when transferred to both retrieval-based and localization-based downstream tasks. 

\noindent \textbf{Acknowledgements.} This paper was partially supported by NSFC (No: 621760\\08) and Shenzhen Science \& Technology Research Program (No:  GXWD20201231\\165807007-20200814115301001).

\clearpage
%
%
\bibliographystyle{splncs04}
\bibliography{egbib}
\end{document}